\documentclass{article}
\PassOptionsToPackage{hyphens}{url} 

\usepackage[main, final]{neurips_2026}

\usepackage[utf8]{inputenc} 
\usepackage[T1]{fontenc}    
\usepackage{hyperref}       
\usepackage{url}            
\usepackage{booktabs}       
\usepackage{array}
\usepackage{amsfonts,amsmath,amsthm,amssymb}       
\usepackage{nicefrac}       
\usepackage{microtype}      
\usepackage{xcolor}         
\usepackage{algorithm}
\usepackage{algpseudocode}
\usepackage{enumitem}
\usepackage{tikz}

\DeclareMathOperator*{\argmax}{arg\,max}
\DeclareMathOperator*{\argmin}{arg\,min}

\newcommand{\bx}{\mathbf{x}}
\newcommand{\be}{\mathbf{e}}
\newcommand{\tQ}{\texttt{Query}}
\newcommand{\LearnHalfspace}{\texttt{LearnHalfspace}}
\newcommand{\codeurl}{\url{https://github.com/nicteh/Learning-Acceptable-Lotteries-Minimizing-Violation-Distances}}

\newtheorem{theorem}{Theorem}[section]
\newtheorem{proposition}[theorem]{Proposition}
\newtheorem{lemma}[theorem]{Lemma}
\newtheorem{corollary}[theorem]{Corollary}
\theoremstyle{definition}

\theoremstyle{remark}

\title{Learning Acceptable Lotteries via Queries: Minimizing Aggregated Violation Distances}

\author{%
  Nicholas Teh \\
  University of Oxford \\
  \texttt{nicholas.teh@cs.ox.ac.uk} \\
  \And
  Paul W.~Goldberg \\
  University of Oxford \\
  \texttt{paul.goldberg@cs.ox.ac.uk} \\
}

\begin{document}

\maketitle

\begin{abstract}
When deploying an AI system, a developer can randomize over several candidate versions, sending each request to a version chosen at random, to balance the requirements of multiple stakeholders. We study a setting in which each of $n$ stakeholders reveals only whether a proposed lottery over $m$ candidates is acceptable, and each stakeholder's acceptable lotteries lie on one side of an unknown hyperplane. While previous work sought a lottery that every stakeholder accepts, such a lottery need not exist when, for instance, safety and helpfulness requirements are in conflict. We therefore minimize a power mean of the stakeholders' violations; since binary answers reveal neither the size of a violation nor how to compare violations across stakeholders, we measure each violation by a normalized distance to the stakeholder's acceptance boundary. We characterize exactly when accept/reject answers determine these distances. When they do, $\mathcal{O}(nm\log(1/\varepsilon))$ queries suffice to learn them, where $\varepsilon$ is the precision of the utilities, and this bound is tight; when they do not, a natural worst-case approach is optimal up to a constant factor, even compared with methods that make unlimited queries. For limited query budgets, we give a sampling algorithm whose number of queries for a given accuracy does not depend on $n$, and show that a deterministic algorithm cannot guarantee a small loss without querying almost all stakeholders. Synthetic experiments illustrate the resulting tradeoff between querying many stakeholders and learning each of them accurately.
\end{abstract}

\section{Introduction}
\label{sec:introduction}

Making a language model less likely to give harmful replies tends to make it refuse more harmless requests \citep{bai2022training,rottger2024xstest}, and different parties care about these two kinds of error to different degrees. Suppose that a developer has a \emph{cautious} version of a chatbot, which produces a harmful reply on $0.1\%$ of risky prompts but refuses $15\%$ of harmless requests, and a \emph{permissive} version, whose rates are $0.9\%$ and $3\%$, respectively. Rather than committing to one version, the developer can route each conversation to a version chosen at random, as is routine in A/B testing \citep{kohavi2020trustworthy}. By linearity of expectation, routing $55\%$ of the conversations to the cautious version yields rates of $0.46\%$ and $9.6\%$, which meet both a safety team's limit of $0.5\%$ on harmful replies and a customer's limit of $10\%$ on refusals, even though neither version meets both limits on its own. Now suppose that a regulator also caps harmful replies at $0.3\%$. Then at least $75\%$ of the conversations must go to the cautious version, whereas the customer's limit allows at most $7/12$ of them to go there, so no mixture satisfies everyone. Yet the developer still has to deploy something. How should it choose a mixture when every mixture leaves some requirement unmet?

We formalize such decisions using \emph{lotteries} over a finite set of alternatives (here, the versions of the system). Following the standard expected utility model \citep{vonNeumannMorgenstern1944,Fishburn1982EU}, each stakeholder, or \emph{agent}, has an unknown score for each alternative and an unknown threshold, and accepts a lottery exactly when its expected score meets the threshold; equivalently, its acceptable lotteries lie on one side of a hyperplane. For requirements on average rates, such as that of harmful replies, this linear model is exact, since the rate of a mixture is the corresponding mixture of rates.

In practice, however, the developer observes neither the scores nor the thresholds. Customers may be unwilling to share their test data, a regulator's requirements may exist only in legal text \citep{euaiact2024}, and a safety team's thresholds may be confidential or hard to articulate. What a stakeholder can readily do is evaluate a concrete proposal and say whether it passes; indeed, evaluations of this kind are central to current AI safety practice, where they determine whether a system may be released \citep{shevlane2023model,weidinger2023sociotechnical}. Following \citet{angluin1988queries}, we call a question of the form ``Would agent $i$ accept lottery $\bx$?''\ a \emph{membership query}. \citet{choo2026learnqueries} showed how to use such queries to decide whether some lottery is acceptable to every agent and, if so, to find one. As our example illustrates, however, requirements may conflict, in which case no lottery is unanimously acceptable, and keeping the current system does not avoid the choice (it is itself one of the alternatives). Which lottery should be chosen when unanimity is impossible, and what can be learned about this choice from accept/reject answers alone?

A natural first idea is to minimize the number of agents who reject the chosen lottery; \citet{teh2026minviol} studied this objective in the same query model. However, the number of rejections does not distinguish between lotteries that violate the same requirement by different amounts. In our example, deploying only the cautious version and routing $75\%$ of the conversations to it are each rejected by the customer alone, but the former misses the customer's limit by five percentage points and the latter by only two. When some requirements concern safety, how severely a requirement is violated matters as much as whether it is violated. Measuring severity is subtle, since the raw shortfall from a threshold has no meaning on its own. Indeed, multiplying an agent's scores and threshold by the same positive constant, or adding the same constant to all of them, leaves every accept/reject answer unchanged but changes the shortfall. Hence raw shortfalls are neither determined by the answers nor comparable across agents (a shortfall of $0.1$ percentage points in harmful replies and one of $0.1$ points in refusals need not be equally serious).

Instead, we measure violation in the space of lotteries. After normalizing the acceptance boundary of each agent, we take the Euclidean distance, within the affine hull of the simplex, from a rejected lottery to this boundary; the distance is zero for accepted lotteries and does not change under the transformations above. We then aggregate the distances of all agents by a power mean with parameter $p\ge 1$: the case $p=1$ measures the average violation, while larger values of $p$ put more weight on the largest violations, which is appropriate when a large violation of a single requirement is unacceptable. Once the acceptance boundaries are known, minimizing this objective is a convex problem. The central question is therefore whether membership queries reveal enough about the boundaries. Classifying lotteries correctly does not suffice, since the objective depends on how far a rejected lottery lies from each normalized boundary.

\subsection{Our Results}

In this paper, we consider $n$ agents and $m$ alternatives. Each agent accepts the lotteries on one side of an unknown hyperplane, and the learner can only ask membership queries. As is standard in query learning, we assume that answers are exact. This assumption is most appropriate when acceptance is decided by an automated evaluation, such as a benchmark with a fixed passing threshold, and is an idealized baseline for human reviewers, who may be uncertain near their boundaries; Section~\ref{subsec:extensions} discusses noisy answers. We describe the model formally in Section~\ref{sec:preliminaries}.

In Section~\ref{sec:full-info}, we begin by asking what an unlimited number of queries can reveal. We show that the answer depends only on how an agent treats the pure alternatives (Theorem~\ref{thm:identification}). If she strictly accepts some pure alternative or rejects exactly one, her normalized boundary is determined by the answers, and $\mathcal{O}(m\log(1/\varepsilon))$ queries suffice to find it, where $\varepsilon$ is the precision of the utilities; this bound is tight in general. Otherwise, the agent is \emph{distance-underdetermined}: the answers determine her acceptable set exactly, but not how fast her violation grows away from it (Figure~\ref{fig:three-cases}). For such an agent, the largest violation consistent with all answers is the distance to her acceptable set (Lemma~\ref{lem:largest-violation}). On the positive side, minimizing the resulting worst-case objective incurs additive loss at most $\sqrt{2}(|B|/n)^{1/p}$, where $B$ is the set of distance-underdetermined agents (Theorem~\ref{thm:worst-case-objective}). On the negative side, no algorithm can guarantee a loss below $(|B|/n)^{1/p}/(12\sqrt{2})$, even with unlimited queries and randomization (Theorem~\ref{thm:ambiguity-lower-bound}). We complement these results with two consequences of convex duality that hold once the boundaries are known: an optimal lottery gives each violated agent a weight that increases with her violation and uses only alternatives that maximize the weighted score (Theorem~\ref{thm:dual-finite-p}), and some optimal lottery uses at most $n+1$ alternatives (Theorem~\ref{thm:few-alternatives}).

Next, in Section~\ref{sec:budget}, we turn to limited query budgets. Our algorithm samples $k$ agents, learns their boundaries, and minimizes the objective of the sample. We show that its loss splits into a boundary-learning term and a sampling term that depends on neither $n$ nor $m$ (Theorem~\ref{thm:alg_error_bound}). When every agent strictly accepts some pure alternative, the algorithm achieves additive error $\alpha$ with probability at least $1-\delta$ using $\mathcal{O}(m\log(1/\varepsilon)\log(2/\delta)\,\alpha^{-2p})$ queries for each fixed $p$, independently of $n$ (Corollary~\ref{cor:main_guarantee}). On the other hand, a deterministic algorithm that queries only $K<n$ agents can suffer additive loss $\Omega(((n-K)/n)^{1/p})$ (Theorem~\ref{thm:unqueried-agents}), so it must query almost all agents in order to guarantee a small loss. We also extend our main guarantees to answers that are flipped independently at random (Proposition~\ref{prop:noise}) and to agents with different weights.

Finally, in Section~\ref{sec:experiments}, we use synthetic experiments as controlled checks of the theory. They illustrate how a fixed budget should be split between querying many agents and learning each agent accurately, and how larger values of $p$ move the optimum toward reducing the largest violations.

\subsection{Related Work}

The paper most closely related to ours is that of \citet{choo2026learnqueries}, who introduced this query model for lotteries and showed how to decide whether some lottery is acceptable to all agents and, if so, how to find one. For this purpose, it suffices to learn the acceptable set of each agent. A major difference between our work and theirs is that we also compare lotteries that some agents reject, which requires the normalized distance to each boundary; accept/reject answers do not always determine this distance, even when they determine the acceptable set. For learning acceptable sets, we use their single-agent procedure (Lemma~\ref{lem:learn-halfspace}). In a companion paper, \citet{teh2026minviol} studied the same query model with the objective of minimizing the number of rejecting agents, which depends only on the acceptable sets but is APX-hard to optimize even when all boundaries are known; by contrast, our objective is convex once the boundaries are known. The two objectives are complementary: ours distinguishes slight violations from severe ones, whereas counting rejections is appropriate for requirements that are simply met or not. Our model combines the expected utility view of lotteries \citep{vonNeumannMorgenstern1944} with a threshold view of acceptability \citep{simon1955behavioral}, but our results depend only on the induced accept/reject answers; Appendix~\ref{app:halfspace-assumption} discusses this assumption.

Our motivation is related to recent work on aligning AI systems with the diverse preferences of the people they affect. \citet{sorensen2024roadmap} argued that AI systems should represent a plurality of human values rather than a single aggregate, and \citet{conitzer2024socialchoice} argued that social choice theory should guide how diverse human feedback is combined. In a similar vein, \citet{neoh2026ensemble} selected a small ensemble of AI systems using binary or pairwise feedback while saving on costly evaluations. Our model is narrower: agents only reveal whether a randomized deployment is acceptable to them, and the developer must still choose a deployment when none is acceptable to all.

Randomization over outcomes is central in probabilistic social choice and randomized mechanism design \citep{gibbard1977manipulation,brandt2017rolling,brandt2016handbook}. Closer in spirit are fair mixing and models with dichotomous preferences, where agents approve or disapprove deterministic outcomes and the decision is a lottery over them \citep{bogomolnaia2005collective,aziz2019fair}, as well as portioning and budget aggregation, where a divisible resource is split among projects based on voters' ordinal or cardinal reports \citep{freeman2021truthful,airiau2023portioning,freeman2024project,elkind2026settling}; we refer to the survey by \citet{suksompong2026voting} for a recent overview of voting over divisible resources. These works typically assume richer preference reports and study axiomatic or strategic properties of aggregation rules, whereas our learner only receives accept/reject answers and must recover enough geometry to compare violations.

Lotteries also appear in constrained machine learning, where randomized classifiers are used to satisfy fairness, error-rate, or resource constraints in expectation \citep{AgarwalEtAl2018,CotterEtAl2019,zafar2019fairness,agarwal2022power}. There, the constraints are usually specified by the designer or estimated from labeled data, whereas each of our constraints belongs to an agent and can only be accessed through accept/reject answers.

Methodologically, our work lies between preference elicitation and query learning \citep{blum2004preference,braziunas2008elicitation}. Learning with membership queries goes back to \citet{angluin1988queries} and is a standard active learning setting \citep{settles2009active}; recent work has studied membership query algorithms for halfspaces \citep{alabdulmohsin2015efficient,kelner2020learning,diakonikolas2024active}. These works aim for classification accuracy, whereas optimizing distances to acceptance boundaries requires the normalized hyperplanes themselves, which binary labels may not reveal.

Finally, once unanimity is impossible, our objective is related to soft constraints, least-violation optimization, and inequality-aware aggregation \citep{boyd2004convex,dai2021optimization,chen2023guide,atkinson1970measurement}. Power means interpolate between average-case and worst-case criteria and have recently been studied as learnable social welfare functions \citep{pardeshi2024learning}; we apply them to normalized distances, so that $p$ interpolates between the average and the maximum violation. Appendix~\ref{app:related-work} discusses further connections.

\section{Preliminaries}
\label{sec:preliminaries}

Let $N=[n]$ be the set of agents and $S=\{s_1,\dots,s_m\}$ the set of alternatives, where $m\ge 2$ and we write $[z]:=\{1,\dots,z\}$ for each positive integer $z$. A lottery is a vector $\bx\in\Delta(S):=\{\bx\in\mathbb{R}^m_{\ge 0}:\sum_{j=1}^m x_j=1\}$, where $x_j$ is the probability of alternative $s_j$, and $\be_j$ denotes the pure lottery on $s_j$. We write $\langle\cdot,\cdot\rangle$ and $\|\cdot\|$ for the Euclidean inner product and norm, and $\mathrm{dist}(\bx,Y):=\inf_{\mathbf{y}\in Y}\|\bx-\mathbf{y}\|$ for a set $Y\subseteq\mathbb{R}^m$. Note that $\|\bx\|\le 1$ for every lottery $\bx$, and any two lotteries are at distance at most $\sqrt{2}$.

Each agent $i\in N$ has a utility vector $u_i\in[0,1]^m$ and a threshold $\tau_i\in(0,1]$, and accepts a lottery $\bx$ if and only if $U_i(\bx):=\langle u_i,\bx\rangle\ge\tau_i$. Her \emph{acceptable set} is $A_i:=\{\bx\in\Delta(S):U_i(\bx)\ge\tau_i\}$, and we let $M_i:=\{j\in[m]:U_i(\be_j)\ge\tau_i\}$ be the set of indices of the pure alternatives that she accepts. We assume that each agent accepts some lottery and rejects some lottery; since a linear function on $\Delta(S)$ attains its maximum and minimum at pure lotteries, this means that $\varnothing\neq M_i\neq[m]$. Appendix~\ref{app:halfspace-assumption} shows that every such linear threshold rule over lotteries can be written in this form.

\textbf{Membership queries.} A membership query asks whether agent $i$ accepts lottery $\bx$; its answer is $\tQ(i,\bx)=1$ if $U_i(\bx)\ge\tau_i$ and $\tQ(i,\bx)=0$ otherwise. Queries may be adaptive. Let $T_i$ be the number of queries to agent $i$, $Q:=\sum_{i\in N}T_i$ the total number of queries, and $K:=|\{i\in N:T_i>0\}|$ the number of agents who receive at least one query. This is the standard membership query model \citep{angluin1988queries,KearnsVazirani1994}. As shown by \citet{choo2026learnqueries}, exact recovery with finitely many queries requires the parameters to have finite precision. Following them, we call an instance \emph{$\varepsilon$-quantized}, where $\varepsilon\in(0,1/2]$ and $1/\varepsilon$ is an integer, if $u_i(s_j)\in\{0,\varepsilon,2\varepsilon,\dots,1\}$ and $\tau_i\in\{\varepsilon,2\varepsilon,\dots,1\}$ for all $i\in N$ and $j\in[m]$. We use this assumption only in our query bounds.

\textbf{Normalized violations.} Accept/reject answers do not change if we add the same constant to the utilities and the threshold of an agent (since the probabilities of a lottery sum to one), or if we multiply them by the same positive number. We remove these two degrees of freedom before comparing agents. Let $\bar u_i:=\frac1m\sum_{j=1}^m u_i(s_j)$ and $w_i:=u_i-\bar u_i\mathbf{1}$, where $\mathbf{1}$ is the all-ones vector; $w_i\neq 0$ because $\varnothing\neq M_i\neq[m]$. The \emph{normalized acceptance boundary} of agent $i$ is the pair $(\hat u_i,\hat\tau_i):=(w_i/\|w_i\|,\,(\tau_i-\bar u_i)/\|w_i\|)$. It satisfies $\|\hat u_i\|=1$ and $\sum_{j=1}^m\hat u_i(s_j)=0$, and agent $i$ accepts $\bx$ if and only if $\langle\hat u_i,\bx\rangle\ge\hat\tau_i$. The \emph{violation} of agent $i$ at $\bx$ is
\[
v_i(\bx):=\max\{0,\hat\tau_i-\langle\hat u_i,\bx\rangle\}=\frac{\max\{0,\tau_i-U_i(\bx)\}}{\|w_i\|},
\]
so $v_i(\bx)=0$ exactly when agent $i$ accepts $\bx$. Geometrically, let $H_i:=\{\mathbf{y}\in\mathbb{R}^m:\sum_{j=1}^m y_j=1\text{ and }\langle\hat u_i,\mathbf{y}\rangle=\hat\tau_i\}$ be the acceptance boundary of agent $i$ within the affine hull of $\Delta(S)$. Since $\hat u_i$ is a unit vector whose coordinates sum to zero, $v_i(\bx)=\mathrm{dist}(\bx,H_i)$ for every lottery $\bx$ that agent $i$ rejects. This is the distance to a hyperplane, which can be smaller than $\mathrm{dist}(\bx,A_i)$. Moreover, since $H_i$ contains a lottery (on the segment between a lottery that agent $i$ accepts and one that she rejects), we have $v_i(\bx)\le\sqrt{2}$ for every lottery $\bx$. Appendix~\ref{app:normalization-details} gives the details and an example.

We aggregate violations by the power mean $V_p(\bx):=\left(\frac1n\sum_{i\in N}v_i(\bx)^p\right)^{1/p}$ for $p\in[1,\infty)$, and let $V_\infty(\bx):=\max_{i\in N}v_i(\bx)$. For finite $p$, it is often convenient to minimize $F(\bx):=V_p(\bx)^p$ instead, which has the same minimizers. Since each $v_i$ is the maximum of two affine functions, it is convex, and so are $F$ and $V_p$. Hence, once the normalized boundaries are known, minimizing $V_p$ over $\Delta(S)$ is a convex optimization problem.

\section{Learning from Unlimited Queries} \label{sec:full-info}

We begin by assuming that every agent can be queried as often as needed.

\subsection{What Membership Queries Reveal} \label{sec:identification}

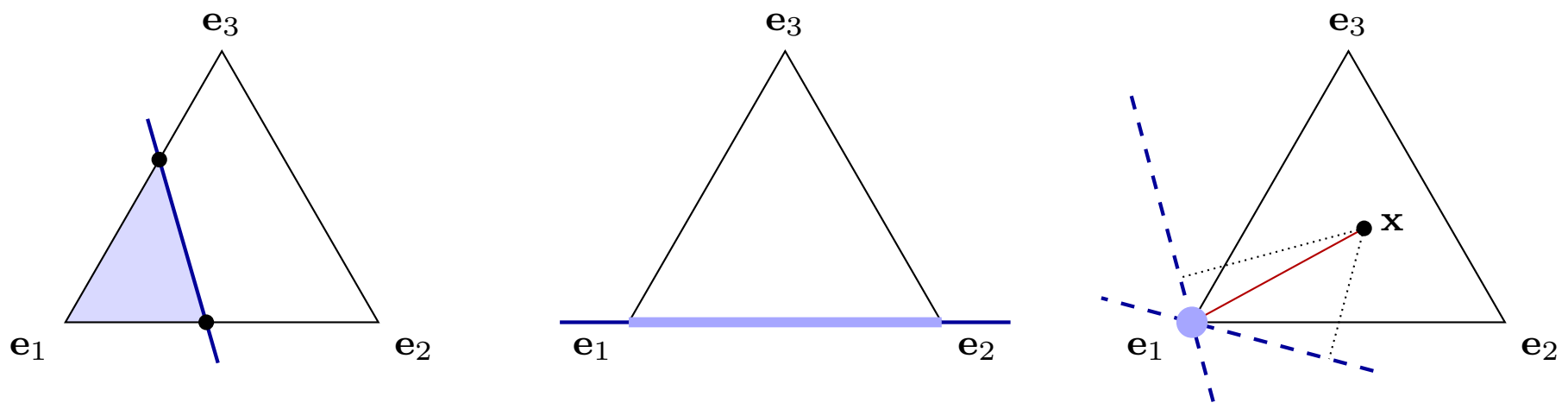
\begin{figure}[t]
\centering
\begin{tikzpicture}[scale=2.4, every node/.style={font=\footnotesize}]
  \begin{scope}
    \fill[blue!15] (0,0) -- (0.45,0) -- (0.3,0.5196) -- cycle;
    \draw (0,0) -- (1,0) -- (0.5,0.866) -- cycle;
    \draw[thick, blue!60!black] (0.4875,-0.1299) -- (0.2625,0.6495);
    \fill (0.45,0) circle (0.025);
    \fill (0.3,0.5196) circle (0.025);
    \node[below left] at (0,0) {$\be_1$};
    \node[below right] at (1,0) {$\be_2$};
    \node[above] at (0.5,0.866) {$\be_3$};
    \node at (0.5,-0.33) {(a) Case 1};
  \end{scope}
  \begin{scope}[xshift=1.8cm]
    \draw (0,0) -- (1,0) -- (0.5,0.866) -- cycle;
    \draw[thick, blue!60!black] (-0.22,0) -- (1.22,0);
    \draw[line width=2.2pt, blue!35] (0,0) -- (1,0);
    \node[below left] at (0,0) {$\be_1$};
    \node[below right] at (1,0) {$\be_2$};
    \node[above] at (0.5,0.866) {$\be_3$};
    \node at (0.5,-0.33) {(b) Case 2};
  \end{scope}
  \begin{scope}[xshift=3.6cm]
    \draw (0,0) -- (1,0) -- (0.5,0.866) -- cycle;
    \draw[dashed, thick, blue!60!black] (0.0906,-0.3381) -- (-0.1941,0.7244);
    \draw[dashed, thick, blue!60!black] (0.5796,-0.1553) -- (-0.2898,0.0776);
    \draw[densely dotted] (0.55,0.30) -- (-0.0381,0.1424);
    \draw[densely dotted] (0.55,0.30) -- (0.4381,-0.1174);
    \draw[red!70!black] (0.55,0.30) -- (0,0);
    \fill[blue!35] (0,0) circle (0.05);
    \fill (0.55,0.30) circle (0.025);
    \node[right] at (0.55,0.32) {$\bx$};
    \node[below left] at (-0.03,0) {$\be_1$};
    \node[below right] at (1,0) {$\be_2$};
    \node[above] at (0.5,0.866) {$\be_3$};
    \node at (0.5,-0.33) {(c) Case 3};
  \end{scope}
\end{tikzpicture}
\caption{The three cases of Section~\ref{sec:identification} for $m=3$, drawn in the plane containing the simplex. Acceptable sets are light blue and boundaries dark blue. In (a), searching along the edges from $\be_1$ finds two points on the boundary. In (c), the agent accepts only $\be_1$, and each dashed line gives the same answer to every query; the distance from $\bx$ to the boundary (dotted) depends on the line, and its largest value is the distance from $\bx$ to $\be_1$ (red), as shown in Lemma~\ref{lem:largest-violation}.}
\label{fig:three-cases}
\end{figure}

Consider a single agent $i$. Querying the $m$ pure lotteries reveals the set $M_i$. Since $U_i$ is linear, agent $i$ accepts some lottery with a positive margin if and only if $U_i(\be_j)>\tau_i$ for some $j\in[m]$. This leads to three cases, illustrated in Figure~\ref{fig:three-cases} for $m=3$.
\begin{enumerate}[leftmargin=*,itemsep=0pt,topsep=2pt]
    \item Agent $i$ strictly accepts some pure alternative. Then searching along edges of the simplex between accepted and rejected pure lotteries yields $m-1$ affinely independent points on her boundary, which determine her normalized boundary.
    \item Every pure alternative that agent $i$ accepts lies on her boundary, and she rejects exactly one pure alternative. Then the $m-1$ accepted pure lotteries span her boundary, which is again determined.
    \item Every pure alternative that agent $i$ accepts lies on her boundary, and she rejects at least two pure alternatives. Then her acceptable set $A_i$ is the face of the simplex spanned by the accepted pure lotteries. Many different normalized boundaries give this acceptable set, and hence the same answer to every query, but different violations outside $A_i$.
\end{enumerate}
We call the agents in case~3 \emph{distance-underdetermined}, and let $B:=\{i\in N:U_i(\be_j)=\tau_i\text{ for all }j\in M_i\text{, and }|M_i|\le m-2\}$ be the set of such agents. In the ambiguity statements below, consistency ranges over all profiles in our model, without the quantization restriction used for finite-query bounds.

\begin{theorem} \label{thm:identification}
    There is a deterministic algorithm that, on every $\varepsilon$-quantized instance, makes $\mathcal{O}(nm\log(1/\varepsilon))$ membership queries and finds the normalized boundary $(\hat u_i,\hat\tau_i)$ of every agent $i\notin B$ and the acceptable set $A_i$ of every agent $i\in B$. For each agent $i\in B$, infinitely many normalized boundaries give the same answer as agent $i$ to every membership query. Moreover, if $m\ge 3$, every deterministic algorithm that finds the normalized boundaries of all agents makes $\Omega(nm\log(1/\varepsilon))$ queries on some $\varepsilon$-quantized instance with $B=\varnothing$.
\end{theorem}

In case~1, our algorithm follows the single-agent procedure of \citet{choo2026learnqueries}, who used it to learn acceptable sets. Our contribution in Theorem~\ref{thm:identification} is to determine exactly when the answers also fix the normalized boundary (a question that does not arise when one only seeks a unanimously acceptable lottery). The lower bound follows by counting: there are $(1/\varepsilon)^{\Omega(m)}$ $\varepsilon$-quantized agents with different normalized boundaries, and each query has only two possible answers. The lower bound concerns recovering the boundaries; whether minimizing $V_p$ exactly requires as many queries is open.

\subsection{Agents with Undetermined Distances} \label{sec:ambiguity}

Since queries cannot reveal how fast the violation of an agent in $B$ grows outside her acceptable set, a natural approach is to plan for the largest violation that is consistent with her answers. Our next lemma shows that this largest violation has a simple form. Intuitively, a consistent boundary can be rotated around the face $A_i$, and the violation at $\bx$ is largest when the boundary is perpendicular to the segment from $\bx$ to its closest point in $A_i$, as in Figure~\ref{fig:three-cases}(c).

\begin{lemma} \label{lem:largest-violation}
    Let $i\in B$. The violations of the normalized boundaries that give the same answer as agent $i$ to every membership query are exactly the functions $\bx\mapsto\sum_{j\notin M_i}\delta_jx_j$ on $\Delta(S)$, where $(\delta_j)_{j\notin M_i}$ ranges over all positive numbers with $\sum_{j\notin M_i}\delta_j^2-\frac1m\big(\sum_{j\notin M_i}\delta_j\big)^2=1$. Moreover, for every $\bx\in\Delta(S)$, the largest violation at $\bx$ among these boundaries is $\mathrm{dist}(\bx,A_i)$, and $\max_{\bx\in\Delta(S)}\mathrm{dist}(\bx,A_i)=\sqrt{1+1/|M_i|}$.
\end{lemma}

Replacing the unknown violation of each agent in $B$ by its largest possible value leads to the objective
\[
\bar F(\bx):=\frac1n\sum_{i\notin B}v_i(\bx)^p+\frac1n\sum_{i\in B}\mathrm{dist}(\bx,A_i)^p.
\]
By Lemma~\ref{lem:largest-violation}, $\bar F(\bx)$ is exactly the largest value of $V_p(\bx)^p$ over all profiles that give the same answers as the true profile to every query. Moreover, $\bar F$ is convex, and it is known after the queries of Theorem~\ref{thm:identification}. The next theorem bounds the loss from minimizing $\bar F$ instead of $V_p$. Since $\bar F\ge F$ and the two functions only differ in the terms of the agents in $B$, this loss is controlled by these terms at an optimal lottery.

\begin{theorem} \label{thm:worst-case-objective}
    Fix $p\in[1,\infty)$, let $\bar\bx$ be a minimizer of $\bar F$ over $\Delta(S)$, and let $\bx^*$ be a minimizer of $V_p$ over $\Delta(S)$. Then
    \[
    V_p(\bar\bx)^p-V_p(\bx^*)^p\le\frac1n\sum_{i\in B}\left(\mathrm{dist}(\bx^*,A_i)^p-v_i(\bx^*)^p\right).
    \]
    In particular, $V_p(\bar\bx)\le V_p(\bx^*)+\sqrt{2}\,(|B|/n)^{1/p}$.
\end{theorem}

The first bound depends on the instance. It shows that $\bar\bx$ is optimal whenever some minimizer of $V_p$ is accepted by every agent in $B$, and that the loss is small whenever such a minimizer is close to their acceptable sets. The second bound holds for every instance; by Lemma~\ref{lem:largest-violation}, its factor $\sqrt{2}$ can be replaced by $\max_{i\in B}\sqrt{1+1/|M_i|}$, which is close to $1$ when every agent in $B$ accepts many pure alternatives. Our next result shows that the dependence on $(|B|/n)^{1/p}$ cannot be avoided by any method that only uses membership queries.

\begin{theorem} \label{thm:ambiguity-lower-bound}
    Fix $p\in[1,\infty)$, and let $n$ and $b$ be integers such that $b$ is even and $2\le b\le n$. There exist two profiles with $m=3$ and $|B|=b$, in which all utilities lie in $\{0,1/2,1\}$ and all thresholds are $1$, such that the two profiles give the same answer to every membership query and, for every lottery $\bx$, the additive losses $V_p(\bx)-\min_{\mathbf{y}\in\Delta(S)}V_p(\mathbf{y})$ in the two profiles sum to at least $\frac{1}{6\sqrt{2}}(b/n)^{1/p}$. Consequently, every algorithm that only uses membership queries, even with unlimited queries and randomization, has expected additive loss at least $\frac{1}{12\sqrt{2}}(b/n)^{1/p}$ in one of the two profiles.
\end{theorem}

Together with Theorem~\ref{thm:worst-case-objective}, this determines the worst-case loss caused by distance-underdetermined agents up to a factor of $24$ when $|B|$ is even. The profiles in the proof are of the kind shown in Figure~\ref{fig:three-cases}(c): in both profiles, one agent accepts only $s_1$ and another accepts only $s_2$, but the rates at which their violations grow toward the other alternatives are swapped, so the optimal lottery leans toward $s_1$ in one profile and toward $s_2$ in the other. Since the two profiles are mirror images under swapping $s_1$ and $s_2$, a lottery that is nearly optimal in one profile is far from optimal in the other.

If $B=\varnothing$, then $\bar F=F$, and Theorems~\ref{thm:identification} and~\ref{thm:worst-case-objective} yield exact optimization.

\begin{corollary} \label{cor:exact}
    Fix $p\in[1,\infty)$. There is a deterministic algorithm that, on every $\varepsilon$-quantized instance with $B=\varnothing$ (for example, one in which every agent strictly accepts some pure alternative) makes $\mathcal{O}(nm\log(1/\varepsilon))$ queries and outputs a minimizer of $V_p$ over $\Delta(S)$.
\end{corollary}

\subsection{Structure of Optimal Lotteries} \label{sec:structure}

We now assume that the normalized boundaries are known and describe optimal lotteries. While both results in this subsection follow from standard convex analysis, we include them because they explain how $p$ determines which agents drive the optimum and show that optimal lotteries can be simple. For $1<p<\infty$, the objective $F$ is differentiable, and its optimality conditions take the following form.

\begin{theorem} \label{thm:dual-finite-p}
    Fix $p\in(1,\infty)$. A lottery $\bx^*$ minimizes $V_p$ over $\Delta(S)$ if and only if every alternative $s_j$ with $x_j^*>0$ maximizes the weighted score $\sum_{i\in N}v_i(\bx^*)^{p-1}\,\hat u_i(s_j)$ over all alternatives.
\end{theorem}

In other words, at an optimum, each agent receives a weight $v_i(\bx^*)^{p-1}$ that increases with her violation, agents who accept $\bx^*$ receive weight zero, and the lottery only uses alternatives that are best for the resulting weighted sum of normalized utilities. Larger values of $p$ give relatively more weight to agents with larger violations. Appendix~\ref{app:dual} shows that these weights, multiplied by $p$, form the unique optimal solution of a dual problem. An analogous description holds for $p=1$ and $p=\infty$, with weights given by an optimal dual solution: for $p=1$, every violated agent has weight $1$, and for $p=\infty$, only the most violated agents can have positive weight (Appendix~\ref{app:endpoint-kkt}).

Theorem~\ref{thm:dual-finite-p} identifies which alternatives an optimal lottery may use, but not how many it needs. We say that a lottery $\bx$ \emph{uses} alternative $s_j$ if $x_j>0$. Since $V_p(\bx)$ depends on $\bx$ only through the $n$ numbers $\langle\hat u_i,\bx\rangle$, Carath\'eodory's theorem yields the following result.

\begin{theorem} \label{thm:few-alternatives}
    Fix $p\in[1,\infty)$. For every minimizer $\bx^*$ of $V_p$ over $\Delta(S)$, there is a lottery $\bar\bx$ that uses at most $n+1$ alternatives, uses only alternatives that $\bx^*$ uses, and satisfies $v_i(\bar\bx)=v_i(\bx^*)$ for every $i\in N$. In particular, $\bar\bx$ also minimizes $V_p$.
\end{theorem}

Allowing a small error reduces the number of alternatives further: Appendix~\ref{app:sparse-lotteries} shows that some lottery that uses at most $\lceil\alpha^{-2}\rceil$ alternatives is within additive $\alpha$ of the optimum, and that for $p\ge 2$, the Frank--Wolfe method approaches the optimum while its $T$-th iterate uses at most $T+1$ alternatives.

\section{Limited Query Budgets} \label{sec:budget}

We now ask what can be guaranteed with a total budget of $Q$ queries. We focus on instances in which no agent is distance-underdetermined, so that the difficulty lies in how to spend the queries rather than in the ambiguity of Section~\ref{sec:ambiguity}. A limited budget creates a breadth-depth tradeoff: querying more agents gives a better picture of the population, but leaves fewer queries for each of them. Relative to the optimum, three kinds of error arise: \emph{boundary-learning error}, since each queried agent's boundary is learned only approximately; \emph{sampling error}, since we optimize over a sample of agents; and the error from agents who receive no queries. The results in this section separate these effects.

\subsection{A Sampling Algorithm} \label{subsec:budgeted-strategy}

Algorithm~\ref{alg:budgeted-fixed-p} samples $k$ agents uniformly at random with replacement, learns the normalized boundary of each sampled agent up to accuracy $\eta$, and minimizes the objective of the sample. It may use any single-agent routine $\LearnHalfspace(i,\eta)$ that, for every agent $i$, makes at most $T(\eta)$ queries to agent $i$ and returns a pair $(\tilde u,\tilde\tau)$ with $\|\tilde u\|=1$, $\sum_{j=1}^m\tilde u(s_j)=0$, $\|\tilde u-\hat u_i\|\le\eta$, and $|\tilde\tau-\hat\tau_i|\le\eta$. The parameters $k$ and $\eta$ are chosen so that $kT(\eta)\le Q$: a larger $k$ gives a more representative sample but forces coarser estimates, while a smaller $k$ gives more accurate estimates of a less representative sample. The last step of the algorithm is a convex optimization problem.

\begin{algorithm}[t]
\caption{Budgeted query strategy for fixed $p$}
\label{alg:budgeted-fixed-p}
\textbf{Input:} Total query budget $Q\in\mathbb{Z}_{>0}$, confidence parameter $\delta\in(0,1)$, and $p\in[1,\infty)$. \\
\textbf{Output:} A lottery $\hat \bx\in\Delta(S)$.
\begin{algorithmic}[1]
\State Choose $k \in \mathbb{Z}_{>0}$ and $\eta \in [0,\tfrac12)$, possibly depending on $Q$, $\delta$, and $p$, such that $kT(\eta)\le Q$.
\State Sample $i_1,\dots,i_k$ independently and uniformly at random from $N$.
\For{$t=1$ to $k$}
  \State $(\tilde u_t,\tilde \tau_t)\leftarrow \LearnHalfspace(i_t,\eta)$, and let $\tilde v_t(\bx):=\max\{0,\tilde \tau_t-\langle \tilde u_t,\bx\rangle\}$ for $\bx\in \Delta(S)$.
\EndFor
\State \Return any $\hat{\bx} \in \argmin_{\bx\in\Delta(S)} \frac{1}{k}\sum_{t=1}^k \tilde v_t(\bx)^p$.
\label{line:alg:strategy_optimizer}
\end{algorithmic}
\end{algorithm}

When an agent strictly accepts some pure alternative, her boundary can be learned exactly.

\begin{lemma}[\citealp{choo2026learnqueries}]\label{lem:learn-halfspace}
    Consider an $\varepsilon$-quantized instance, and let $i\in N$ be an agent with $U_i(\be_j)>\tau_i$ for some $j\in[m]$. There is a deterministic algorithm that makes $\mathcal{O}(m\log(1/\varepsilon))$ queries to agent $i$ and returns $(\hat u_i,\hat\tau_i)$.
\end{lemma}

The procedure of \citet{choo2026learnqueries} returns a linear inequality that defines $A_i$ on $\Delta(S)$. Since agent $i$ is in case~1 of Section~\ref{sec:identification}, normalizing this inequality as in Section~\ref{sec:preliminaries} gives $(\hat u_i,\hat\tau_i)$; the procedure is also case~1 in the proof of Theorem~\ref{thm:identification}. With this routine, we can take $\eta=0$.

Our main result in this section bounds the loss of Algorithm~\ref{alg:budgeted-fixed-p} for any routine with the guarantee above. Its proof compares the true objective $F$, the objective of the sampled agents with their true boundaries, and the objective that the algorithm minimizes, and bounds their differences uniformly over $\Delta(S)$.

\begin{theorem}\label{thm:alg_error_bound}
    Algorithm~\ref{alg:budgeted-fixed-p} makes at most $Q$ membership queries, and with probability at least $1-\delta$ over the draw of $i_1,\dots,i_k$, its output $\hat\bx$ satisfies
    \[
    V_p(\hat{\bx})\le \min_{\bx\in\Delta(S)}V_p(\bx)
    +\min\left\{
    \left(4p\eta(\sqrt{2}+2\eta)^{p-1}+2\rho_k(\delta)\right)^{1/p},
    \;
    4\eta + 2\rho_k(\delta)^{1/p}
    \right\},
    \]
    where $\rho_k(\delta):= 2p(\sqrt{2})^{p-1}/\sqrt{k}+(\sqrt{2})^p\sqrt{\log(2/\delta)/(2k)}$.
\end{theorem}

The term $\rho_k(\delta)$ is the sampling error. It depends on neither $n$ nor $m$; the number of alternatives only enters through the cost $T(\eta)$ of learning one agent. The two expressions in the minimum come from comparing the learned and true objectives at the level of $F=V_p^p$ and of $V_p$, respectively; the second is smaller when the boundary-learning error dominates. With the exact routine of Lemma~\ref{lem:learn-halfspace}, the boundary-learning error vanishes, and a suitable choice of $k$ gives the following guarantee.

\begin{corollary} \label{cor:main_guarantee}
    Fix $p\in[1,\infty)$. There is a constant $C_p>0$, depending only on $p$, such that for all $\alpha>0$ and $\delta\in(0,1)$, there is a randomized algorithm that, on every $\varepsilon$-quantized instance in which every agent strictly accepts some pure alternative, makes at most $C_p\, m\log(1/\varepsilon)\,(1+\log(2/\delta))\,\alpha^{-2p}$ queries and returns a lottery $\hat{\bx}$ such that $V_p(\hat{\bx}) \le \min_{\bx\in\Delta(S)} V_p(\bx) + \alpha$ with probability at least $1-\delta$.
\end{corollary}

The number of queries does not depend on $n$. The exponent $2p$ arises because the sampling error in $F$ decreases at rate $k^{-1/2}$, while an error of $\alpha$ in $V_p=F^{1/p}$ requires an error of order $\alpha^p$ in $F$.

\subsection{Error from Unqueried Agents} \label{subsec:unqueried-agents}

Algorithm~\ref{alg:budgeted-fixed-p} chooses the queried agents at random, and Theorem~\ref{thm:alg_error_bound} shows that the loss due to the agents it does not query is captured by the sampling error $\rho_k(\delta)$. Our next result shows that a deterministic algorithm cannot leave agents unqueried in this way, even if it learns the queried agents exactly. The idea is that the preferences of an agent who is never queried can be chosen after the algorithm has fixed its output, so that this output is far from her acceptable set.

\begin{theorem}\label{thm:unqueried-agents}
    Fix $p \in [1,\infty)$ and $K \in \{0,1,\dots,n-1\}$, and consider any deterministic, possibly adaptive, algorithm that queries at most $K$ distinct agents on every profile. Then there is a profile with $m=3$, $u_i \in \{0,1\}^3$, and $\tau_i = 1$ for all $i \in N$, on which the algorithm outputs a lottery $\hat\bx$ with
    $V_p(\hat \bx) \ge \min_{\bx \in \Delta(S)} V_p(\bx) + \sqrt{3/8}\,((n-K)/n)^{1/p}$. The bound also holds with constant $\sqrt{6}/8$ on a profile with $m=3$, $u_i\in\{0,1\}^3$, and $\tau_i\in\{\tfrac12,\tfrac34\}$ for all $i$, in which every agent strictly accepts some pure alternative.
    In particular, every deterministic algorithm that always outputs a minimizer of $V_p$ must query all $n$ agents on some profile.
\end{theorem}

This bound is tight up to a constant factor: if the exact boundaries of $K$ agents are known, minimizing their contribution to $F$ yields additive loss at most $\sqrt{2}((n-K)/n)^{1/p}$ (Proposition~\ref{prop:coverage_upper}). An analogous phenomenon arises for the number of rejecting agents: for $m\ge 3$, \citet{teh2026minviol} showed that a deterministic algorithm querying at most $K$ distinct agents can be forced to incur $n-K$ more rejections than necessary. Theorem~\ref{thm:unqueried-agents} does not apply to randomized algorithms such as Algorithm~\ref{alg:budgeted-fixed-p}, which choose the queried agents at random. For randomized algorithms, we only obtain a weaker statement: an algorithm that always outputs a minimizer of $V_p$ must, for each agent $i$, query agent $i$ with probability one on some profile (Proposition~\ref{prop:randomized-exact}). Proving a lower bound for randomized algorithms that matches the guarantee of Algorithm~\ref{alg:budgeted-fixed-p} remains open.

\subsection{Noisy Answers and Weighted Agents} \label{subsec:extensions}

When each answer is wrong independently with probability below $1/2$, repeating each query and taking the majority answer reduces the problem to the case of exact answers.

\begin{proposition}\label{prop:noise}
    Let $\nu\in[0,1/2)$ and $\delta\in(0,1)$, and suppose that each answer to a membership query, including each repeated answer to the same query, is flipped independently with probability at most $\nu$. Consider any algorithm that makes at most $L$ queries when answers are exact. Modify it so that it asks each query $r$ times and uses the majority answer, where $r$ is the smallest odd integer that is at least $2\ln(L/\delta)/(1-2\nu)^2$, and so that it stops after $L$ queries, returning an arbitrary output if it has not finished by then. Then the modified algorithm makes at most $rL$ queries, and with probability at least $1-\delta$, every majority answer is correct, so that it behaves exactly as the original algorithm does with exact answers.
\end{proposition}

For example, with the routine of Lemma~\ref{lem:learn-halfspace}, Algorithm~\ref{alg:budgeted-fixed-p} makes $L=\mathcal{O}(km\log(1/\varepsilon))$ queries. Applying Proposition~\ref{prop:noise} and Theorem~\ref{thm:alg_error_bound} with $\delta/2$ in place of $\delta$, we obtain the guarantee of Theorem~\ref{thm:alg_error_bound} with $\rho_k(\delta/2)$ in place of $\rho_k(\delta)$ with probability at least $1-\delta$, using at most $rL$ queries, where $r=\mathcal{O}((1-2\nu)^{-2}\log(2L/\delta))$. Proposition~\ref{prop:noise} does not cover answers that are arbitrary or correlated near the boundary, nor strategic answers. In such cases, exact recovery is impossible without further assumptions, such as a margin around the boundary, but Theorem~\ref{thm:alg_error_bound} still applies to any routine that learns each boundary up to accuracy $\eta$.

Our model gives every agent the same weight, which avoids deciding whose acceptability should count more. In some applications, however, agents have different entitlements, as in weighted fair division \citep{suksompong2022weighted,suksompong2023weighted,montanari2025weighted,teh2026weightedmixed,suksompong2025weightedreview}. For finite $p$, Theorems~\ref{thm:worst-case-objective}, \ref{thm:dual-finite-p}, and~\ref{thm:alg_error_bound} extend to the weighted objective $\big(\sum_{i\in N}a_iv_i(\bx)^p\big)^{1/p}$, where $a_i\ge 0$ and $\sum_{i\in N}a_i=1$. In Theorem~\ref{thm:worst-case-objective}, the term $(|B|/n)^{1/p}$ becomes $(\sum_{i\in B}a_i)^{1/p}$; in Theorem~\ref{thm:dual-finite-p}, the weight of agent $i$ becomes $a_iv_i(\bx^*)^{p-1}$; and if Algorithm~\ref{alg:budgeted-fixed-p} samples each agent $i$ with probability $a_i$, Theorem~\ref{thm:alg_error_bound} holds unchanged (Appendix~\ref{app:weights}).

\section{Experiments}
\label{sec:experiments}

We use synthetic experiments to illustrate two effects predicted by the theory; they are not meant as evidence of performance in deployment. First, a fixed query budget creates a tradeoff between querying many agents and learning each of them accurately (Theorem~\ref{thm:alg_error_bound}). Second, the parameter $p$ trades off the average violation against the largest violations (Theorem~\ref{thm:dual-finite-p}). Our code is available at \codeurl.

For the first effect, we run Algorithm~\ref{alg:budgeted-fixed-p} with $p=2$ on two populations, each with $n=1000$ agents and $m=10$ alternatives. In the \emph{shared-block} population, agents are noisy variations of a common preference pattern. In the \emph{conflicting-blocks} population, five clusters of agents favor disjoint blocks of alternatives, and the smallest cluster has a stricter threshold. Both populations are $\varepsilon$-quantized with $\varepsilon=1/32$, and every agent strictly accepts some pure alternative. The algorithm learns each sampled agent with the procedure of Lemma~\ref{lem:learn-halfspace}, which runs a binary search on each of $m-1$ edges of the simplex. With $109$ queries, these searches recover the boundary exactly; with fewer queries, we stop each search early and fit the boundary through the midpoints of the final search intervals. Since this learner then has no accuracy guarantee, the experiments illustrate the tradeoff described by Theorem~\ref{thm:alg_error_bound} rather than test it. Under a total budget of $Q=3200$ queries, each of the $k$ sampled agents receives $\lfloor Q/k\rfloor$ queries. Appendix~\ref{app:experiments} gives the details.

\begin{figure}[t]
  \centering
  \includegraphics[width=\linewidth]{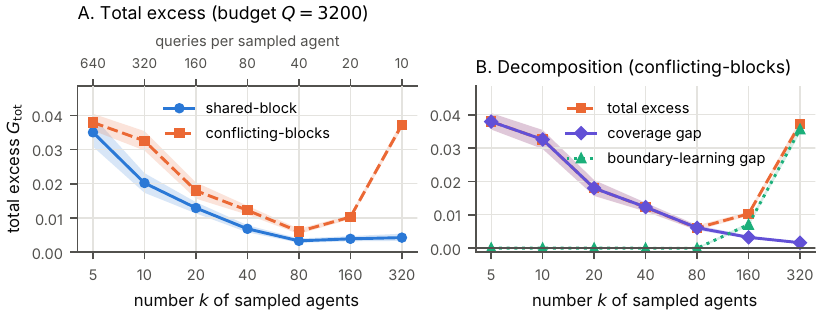}
  \vspace{-0.2in}
  \caption{\textbf{A fixed query budget creates a tradeoff between coverage and accuracy.} Panel A shows the total excess $G_{\mathrm{tot}}$ as the number $k$ of sampled agents varies under a fixed budget of $Q=3200$ queries; the top axis shows the number $\lfloor Q/k\rfloor$ of queries per sampled agent, of which at most $109$ are used. Panel B splits the total excess for the conflicting-blocks population into the coverage gap and the boundary-learning gap. Lines show means over 100 populations, and shaded bands are 95\% bootstrap confidence intervals for the means.}
  \label{fig:query-budget}
  \vspace{-0.1in}
\end{figure}

To separate the two sources of error, let $\bx_{\mathrm{sample}}$ be a lottery that minimizes the objective of the sampled agents with their true boundaries. We split the total excess $G_{\mathrm{tot}}=V_2(\hat\bx)-\min_{\bx}V_2(\bx)$ into the \emph{coverage gap} $G_{\mathrm{cov}}=V_2(\bx_{\mathrm{sample}})-\min_{\bx}V_2(\bx)$ and the \emph{boundary-learning gap} $G_{\mathrm{bdry}}=V_2(\hat\bx)-V_2(\bx_{\mathrm{sample}})$, which play the roles of the sampling and boundary-learning terms in Theorem~\ref{thm:alg_error_bound}. Figure~\ref{fig:query-budget} shows that the budget must be split carefully. For $k\le 20$, every sampled agent receives at least $109$ queries and is learned exactly, so the total excess consists only of the coverage gap; for the conflicting-blocks population, it is $3.80\times 10^{-2}$ at $k=5$. As $k$ grows, the coverage gap shrinks, but each sampled agent is learned less accurately. At $k=320$, each sampled agent receives only $10$ queries, just enough to query the pure lotteries, and the boundary-learning gap is $3.57\times 10^{-2}$, about 96\% of the total excess. Among the tested values of $k$, the smallest mean total excess for the conflicting-blocks population is $6.02\times 10^{-3}$, attained at $k=80$ with $40$ queries per sampled agent and significantly smaller than for every other $k$. For the shared-block population, the smallest mean total excess, $3.30\times 10^{-3}$, is also attained at $k=80$, but the differences from $k=160$ and $k=320$ are within the noise. The best split depends on the population, however: if the thresholds in the conflicting-blocks population lie closer to the middle of each agent's utility range, the coarse boundaries learned from $10$ queries hurt much less, and $k=320$ does as well as $k=80$ (Appendix~\ref{app:experiments}).

Appendix~\ref{app:experiments} also varies $p$ on a structured instance with three clusters. As predicted by Theorem~\ref{thm:dual-finite-p}, larger values of $p$ favor agents with larger violations: moving from $p=1$ to $p=\infty$ reduces the largest violation from $0.391$ to $0.225$, but increases the average violation from $0.047$ to $0.066$.

\section{Conclusion}
\label{sec:conclusion}

In this paper, we have studied how to choose a lottery that minimizes aggregate violation when agents only reveal whether a proposed lottery is acceptable and no lottery need be acceptable to all of them. We showed that accept/reject answers determine the normalized violation of every agent except the distance-underdetermined ones, determined the unavoidable loss caused by these agents up to a constant factor, and analyzed a sampling algorithm whose number of queries does not depend on the number of agents. Overall, our results indicate that binary feedback suffices to optimize the severity of violations, and not only their number (studied by \citet{teh2026minviol}), as long as no agent is distance-underdetermined. However, minimizing aggregate violation is no substitute for safety requirements that must hold without exception; such requirements should be imposed as constraints on the candidate lotteries rather than traded off against other requirements.

A technical question left by our work is to close the constant-factor gap between Theorems~\ref{thm:worst-case-objective} and~\ref{thm:ambiguity-lower-bound}, and to prove a lower bound for randomized algorithms that query few agents, complementing Theorem~\ref{thm:unqueried-agents}. It would also be interesting to extend the limited-budget analysis of Section~\ref{sec:budget} to distance-underdetermined agents, and to determine whether exact optimization requires as many queries as exact recovery of the boundaries (Theorem~\ref{thm:identification}). When additional information is available, such as a prior over the boundaries consistent with the answers, one could replace the worst-case objective $\bar F$ by a less pessimistic one, in the spirit of learning-augmented algorithms \citep{mitzenmacher2021algorithms} (see also Appendix~\ref{app:related-work}). Another direction is to replace the simplex by a general polytope of feasible outcomes; Appendix~\ref{app:polytopes} explains which parts of our analysis extend and why our query algorithm does not. Finally, a study of elicitation from real stakeholders, whose answers may be uncertain or strategic, would test how far the theory of learning from acceptability feedback carries over to practice.

\begin{ack}
This work was funded by the Advanced Research + Invention Agency (ARIA) under the ASPAI project. PWG is also supported by EPSRC Grants EP/X040461/1 and EP/X038548/1.
\end{ack}

\bibliographystyle{plainnat}
\bibliography{abb,bib}

\clearpage

\appendix

\section{Modeling Details and Omitted Discussion}
\label{app:modeling-details}

\subsection{Halfspace Acceptability and Expected Utility}
\label{app:halfspace-assumption}

In this appendix, we clarify the role of the expected utility assumption. Our results use expected utility only through a geometric consequence: the acceptable set of each agent is the intersection of the simplex with a halfspace. Our model should therefore be understood primarily as the assumption that acceptable sets are halfspaces, with expected utility as the standard interpretation for lotteries with known probabilities.

\textbf{Linear threshold rules.}
Consider any acceptability rule of the form $A_i=\{\bx\in\Delta(S):\langle \mathbf{c},\bx\rangle\ge d\}$ with $\mathbf{c}\in\mathbb{R}^m$ and $d\in\mathbb{R}$, and assume that $A_i\neq\varnothing$ and $A_i\neq\Delta(S)$. Since a linear function on the simplex attains its minimum and maximum at pure lotteries, we have $c_{\min} < d \le c_{\max}$, where $c_{\min}$ and $c_{\max}$ are the smallest and largest coordinates of $\mathbf{c}$. Let
\[
        u_i:=\frac{\mathbf{c}-c_{\min}\mathbf{1}}{c_{\max}-c_{\min}},
        \quad
        \tau_i:=\frac{d-c_{\min}}{c_{\max}-c_{\min}}.
\]
Then $u_i\in[0,1]^m$ and $\tau_i\in(0,1]$, and since $\sum_{j=1}^m x_j=1$ for every $\bx\in\Delta(S)$, we have $\langle \mathbf{c},\bx\rangle \ge d \iff \langle u_i,\bx\rangle \ge \tau_i$.
Hence the notation of Section~\ref{sec:preliminaries} is without loss of generality for any nontrivial linear threshold rule; conversely, every pair $(u_i,\tau_i)$ in Section~\ref{sec:preliminaries} induces such a rule. In particular, our results rely neither on interpersonal comparisons of utilities nor on the learner observing utilities; they only rely on the assumption that acceptable sets are halfspaces.

\textbf{Interpretation for randomized deployments.}
The alternatives $s_1,\dots,s_m$ can be candidate classifiers, versions of a foundation model, ranking rules, or allocation policies, and a lottery $\bx\in\Delta(S)$ can represent a randomized deployment, a randomized classifier, or a repeated deployment in which alternative $s_j$ is used with frequency $x_j$. If stakeholder $i$ assigns a score $u_i(s_j)$ to each alternative and accepts the deployments whose expected score is at least $\tau_i$, then acceptability is exactly $\sum_{j=1}^m x_j u_i(s_j)\ge \tau_i$. The score need not be a utility in the psychological sense: it can be, for example, the negative of a rate of harmful replies, the accuracy on a subgroup of users, or any other statistic that is averaged over requests. This interpretation is particularly natural in constrained and fair machine learning, where randomized classifiers and constraints in expectation are standard tools \citep{AgarwalEtAl2018,CotterEtAl2019}. The difference in our setting is that the constraints are not known to the learner and must be learned from queries.

\textbf{Scope of the assumption.}
As is common in learning theory, we fix a hypothesis class (here, the halfspaces of the simplex) and ask what can be identified and optimized from limited labels \citep{ShalevShwartzBenDavid2014}. Our feedback model is weaker than one in which cardinal utilities, welfare values, or rankings are observed: the learner never sees $u_i$ or $\tau_i$, and the normalization in Section~\ref{sec:preliminaries} removes offsets and scales before violations are compared. Some structural restriction is also necessary for query-efficient guarantees. Indeed, with an arbitrary membership oracle on the continuum of lotteries, finitely many queries leave the answers at all other lotteries unconstrained, so no nontrivial guarantee for exact optimization is possible. The halfspace assumption is therefore best read as a choice of model class, one that makes identification, normalization, convex optimization, and finite-query analysis possible.

\textbf{Beyond expected utility.}
We do not claim that all human or institutional preferences over lotteries obey expected utility. If acceptability is genuinely nonlinear in the probabilities, exact recovery of halfspaces and the violation geometry of this paper need not apply, and our results should then be viewed as a baseline, analogous to linear models in supervised learning. Two extensions are natural. First, if answers are flipped independently at random, Proposition~\ref{prop:noise} reduces the problem to the case of exact answers. If answers are noisy in other ways, or only approximately described by halfspaces, Theorem~\ref{thm:alg_error_bound} still shows where the additional error enters: any routine that learns each boundary up to accuracy $\eta$ contributes the corresponding boundary-learning term to the final loss, so an active learning method for noisy labels could replace exact recovery. Second, richer acceptability models, such as intersections of halfspaces, piecewise-linear acceptable regions, or more general convex sets, can be handled by enlarging the model class. This would change the analysis of learnability and query complexity, and the optimality conditions and the bound on the number of alternatives in Section~\ref{sec:structure} need not hold unchanged, but the same template remains natural: sample agents, learn their acceptability rules, and optimize the resulting objective.

\subsection{Additional Related Work}
\label{app:related-work}

The main text discusses the work most closely related to ours. Here, we elaborate on our companion paper and discuss several further connections, including the use of predictions to circumvent worst-case lower bounds.

\textbf{Minimizing the number of rejecting agents.}
\citet{teh2026minviol} considered the same query model but a different objective: instead of measuring how severely each agent is violated, they counted the agents who reject the chosen lottery. They showed that the problem is APX-hard already when all boundaries are known and all utilities are $0$ or $1$, gave exact algorithms under additional structure, and proved guarantees, with a sample size independent of $n$, for optimizing over a uniform sample of agents. They also showed that, for $m\ge 3$, a deterministic algorithm querying at most $K$ distinct agents can be forced to incur $n-K$ more rejections than necessary, which parallels Theorem~\ref{thm:unqueried-agents}. The two objectives differ in two respects. First, the number of rejections depends only on the acceptable sets, which membership queries always determine, so the ambiguity of Section~\ref{sec:ambiguity} does not arise for that objective. Second, our objective is convex once the boundaries are known, so the difficulty in our setting is informational rather than computational. Conceptually, a distance-based objective distinguishes slight violations from severe ones, whereas counting rejections is appropriate for requirements that are simply met or not; indeed, a small aggregate distance can hide many agents who each fall slightly short of their thresholds, which is one reason why requirements that must hold without exception should be imposed as constraints (Section~\ref{sec:conclusion}).

\textbf{Identifiability.}
As in much of learning theory, a structural assumption on the hypothesis class makes finite-query guarantees possible \citep{KearnsVazirani1994, ShalevShwartzBenDavid2014}.
In our setting, the hypothesis attached to each agent is a halfspace over the continuum of lotteries, and this assumption leaves an identifiability issue that is specific to acceptability queries: several pairs of utilities and thresholds induce exactly the same membership oracle.
The normalized acceptance boundary is therefore the object that matters for the decision, and the results in Section~\ref{sec:full-info} characterize when this object is identifiable, when only the acceptable set is identifiable, and how the remaining ambiguity affects the optimization problem.

\textbf{Learning over a population of agents.}
Our algorithm for limited budgets resembles empirical risk minimization over a population of agents: each queried agent contributes a learned violation function, and the returned lottery minimizes the empirical aggregate.
The difference from a standard supervised learning pipeline is that each sampled data point is itself an unknown constraint that must first be elicited through membership queries.
This is why our analysis separates boundary-learning error, sampling error, and the effect of unqueried agents.
The uniform convergence step in the proof of Theorem~\ref{thm:alg_error_bound} uses standard symmetrization arguments from statistical learning theory \citep{KearnsVazirani1994, ShalevShwartzBenDavid2014}, applied to a loss class induced by normalized violations of learned acceptance boundaries.

\textbf{Randomizing over few alternatives.}
Theorem~\ref{thm:few-alternatives} gives a convex-analytic reason why an optimal lottery need not randomize over all alternatives: some optimal lottery uses at most $n+1$ alternatives, even though a lottery may use the entire menu.
Unlike the axiomatic guarantees usually studied for voting or allocation rules, this property is a structural feature of the optimization problem that arises once the acceptance boundaries have been learned.
Besides its usual roles as a fairness or strategic device, randomization here reaches compromises that no single alternative achieves.

\textbf{Predictions in collective decision-making.}
Two of our lower bounds are worst-case statements that additional information could circumvent: Theorem~\ref{thm:ambiguity-lower-bound} holds because accept/reject answers cannot distinguish certain boundaries, and Theorem~\ref{thm:unqueried-agents} holds because a deterministic algorithm learns nothing about the agents it never queries. Learning-augmented algorithms address limits of this kind by using predictions that may be erroneous, with guarantees that improve as the predictions become more accurate yet remain meaningful when they are poor \citep{lykouris2021competitive,purohit2018improving,mitzenmacher2021algorithms}. This framework has been applied to mechanism design \citep{xu2022mechanism,balkanski2023strategyproof,agrawal2024learning}, voting \citep{berger2024learning}, and online fair division, where predictions, together with other relaxations such as recourse and weaker adversaries, help circumvent strong impossibility results \citep{banerjee2022online,banerjee2023proportionally,choo2026approximate,neoh2026online,neoh2026closinggaps,neoh2026frequency,cohen2026budget,cohen2026oblivious,huang2026upgrade,huang2026chores,teh2026housesubsidy}. Closer to our setting, \citet{choo2026learnqueries} showed that advice (for example, a candidate lottery that is likely to be acceptable, or a guess as to which requirements are tight) reduces the number of queries needed to reach unanimity. Algorithm~\ref{alg:budgeted-fixed-p} instead relies on randomization: by choosing at random whom to query, it avoids the lower bound of Theorem~\ref{thm:unqueried-agents}, which applies only to deterministic algorithms. Combining random sampling with predictions about the agents' boundaries, for instance to replace the worst-case objective $\bar F$ of Section~\ref{sec:ambiguity} by a less pessimistic one, is an interesting direction for future work.

\textbf{Uncertainty in collective decisions.}
Our learner improves a collective decision by reducing its own uncertainty about the agents' requirements. In a different model, \citet{aloufheffetz2022better} studied how a benevolent external party can improve the outcome of issue-by-issue majority voting when some agents are confused about which option is better for them, by educating some of the agents, appointing a subset of competent agents to decide on behalf of the group, or guiding the delegation of votes. There, the uncertainty lies with the agents rather than with the decision maker, and the main questions concern the computational complexity of choosing such an intervention rather than the number of queries.

\subsection{Normalization Details and an Example}
\label{app:normalization-details}

This subsection gives the details behind the normalized violation in Section~\ref{sec:preliminaries}. Recall that $\bar u_i=\frac1m\sum_{j=1}^m u_i(s_j)$ and $w_i=u_i-\bar u_i\mathbf{1}$. Then $\sum_{j=1}^m w_i(s_j)=0$, and $\langle w_i,\bx\rangle = \langle u_i,\bx\rangle-\bar u_i$ for every $\bx\in\Delta(S)$. Hence
\[
    \langle u_i,\bx\rangle\ge \tau_i
    \iff
    \langle w_i,\bx\rangle\ge \tau_i-\bar u_i .
\]
Since agent $i$ accepts some pure alternative and rejects another, $u_i$ is not constant, so $w_i\neq 0$. Dividing by $\|w_i\|$ gives $(\hat u_i,\hat\tau_i)$, which satisfies
\[
    \|\hat u_i\|=1,
    \quad
    \sum_{j=1}^m \hat u_i(s_j)=0,
    \quad
    \langle u_i,\bx\rangle\ge\tau_i
    \iff
    \langle \hat u_i,\bx\rangle\ge\hat\tau_i .
\]
For a rejected lottery $\bx$, the violation $v_i(\bx)=\hat\tau_i-\langle\hat u_i,\bx\rangle$ is exactly the Euclidean distance from $\bx$ to $H_i$. Indeed, since the coordinates of $\hat u_i$ sum to zero, $\hat u_i$ is parallel to the affine hull of $\Delta(S)$, and it is orthogonal to every direction within $H_i$. Hence the closest point of $H_i$ to $\bx$ is $\bx+(\hat\tau_i-\langle\hat u_i,\bx\rangle)\hat u_i$, and since $\|\hat u_i\|=1$, its distance from $\bx$ is $\hat\tau_i-\langle\hat u_i,\bx\rangle$.

A simple example shows why this normalization matters. Let $m=2$ and write $\bx=(x_1,1-x_1)$. Consider two agents $(u_a,\tau_a)=((1,0),1/2)$ and $(u_b,\tau_b)=((1,1/2),3/4)$.
Both accept exactly the same lotteries, namely those with $x_1\ge 1/2$. However, their shortfalls are
on different scales: $\max\{0,\tau_a-\langle u_a,\bx\rangle\}=\max\{0,\frac12-x_1\}$, whereas
\[
    \max\{0,\tau_b-\langle u_b,\bx\rangle\}
    =
    \max\Big\{0,\frac34-x_1-\frac12(1-x_1)\Big\}
    =
    \frac12\max\Big\{0,\frac12-x_1\Big\}.
\]
Thus shortfalls would treat two identical acceptability rules differently. After centering and
normalizing, both agents have $\hat u=(1/\sqrt2,-1/\sqrt2)$ and $\hat\tau=0$, and hence the same violation $\sqrt2\max\{0,\frac12-x_1\}$.
For rejected $\bx$, this is precisely the Euclidean distance within the line $x_1+x_2=1$ from $\bx$ to
the boundary point $(1/2,1/2)$.

\subsection{Beyond the Simplex}\label{app:polytopes}

In some applications, the feasible outcomes form a known bounded polytope $P\subseteq\mathbb{R}^m$ other than the simplex (for example, budget allocations with additional upper bounds on individual projects). If each agent accepts the points of $P$ on one side of a hyperplane, violations can be defined as in Section~\ref{sec:preliminaries}, by normalizing the hyperplane within the affine hull of $P$ and measuring Euclidean distances within this affine hull. Several parts of our analysis only use convexity and the diameter of the feasible set, and extend with minor changes. The objective remains convex. The optimality condition of Theorem~\ref{thm:dual-finite-p} becomes the condition that the optimal point maximizes the weighted score $\sum_{i\in N}v_i(\bx^*)^{p-1}\langle\hat u_i,\bx\rangle$ over $\bx\in P$. The analysis of Algorithm~\ref{alg:budgeted-fixed-p} goes through with $\sqrt{2}$ replaced by the diameter of $P$ and with the bound $\|\bx\|\le 1$ replaced by a bound on the norms of points in $P$. Moreover, if the boundary of some agent is not determined by her answers, the distance from a rejected point $\bx$ to any boundary consistent with the answers is still at most $\mathrm{dist}(\bx,A_i)$, because such a boundary separates $\bx$ from $A_i$; hence the argument behind Theorem~\ref{thm:worst-case-objective} gives the same bound with $\sqrt{2}$ replaced by the diameter of $P$. By contrast, our query algorithm in Theorem~\ref{thm:identification} relies on the vertices and edges of the simplex, and characterizing what membership queries reveal for a general polytope requires further work: the vertices of $P$ need not be affinely independent, so the three cases of Section~\ref{sec:identification} no longer cover all possibilities. It would also be interesting to understand how the combinatorial structure of $P$, for instance, when $P$ is a polymatroid, affects these questions.

\section{Omitted Proofs from Section~\ref{sec:full-info}}\label{app:section3-extra}

\subsection{Proof of Theorem~\ref{thm:identification}}\label{app:proof-identification}

We first describe a procedure for a single agent $i$ that makes $\mathcal{O}(m\log(1/\varepsilon))$ queries; running it for every agent gives the claimed algorithm. The high-level idea is to find enough points on the boundary of the agent by searching along edges of the simplex, as illustrated in Figure~\ref{fig:three-cases}.

\textbf{Finding the boundary point on an edge.} Let $\mathbf{y}$ and $\mathbf{z}$ be pure lotteries such that agent $i$ rejects $\mathbf{y}$ and accepts $\mathbf{z}$. Since $U_i$ is linear, agent $i$ accepts $(1-\lambda)\mathbf{y}+\lambda\mathbf{z}$ if and only if $\lambda\ge\lambda^*$, where
\[
\lambda^*:=\frac{\tau_i-U_i(\mathbf{y})}{U_i(\mathbf{z})-U_i(\mathbf{y})}\in(0,1].
\]
Since the instance is $\varepsilon$-quantized, $\lambda^*$ is a fraction whose denominator is at most $1/\varepsilon$, and any two distinct such fractions differ by at least $\varepsilon^2$. Hence binary search on $\lambda$ finds $\lambda^*$, and thus the boundary point $(1-\lambda^*)\mathbf{y}+\lambda^*\mathbf{z}$, with $\mathcal{O}(\log(1/\varepsilon))$ queries.

\textbf{The procedure.} First, query all $m$ pure lotteries; this reveals $M_i$, which is nonempty and different from $[m]$. Fix any $b\notin M_i$. For each $j\in M_i$, find the value $\lambda_j^*$ for the edge from $\be_b$ to $\be_j$. Note that $\lambda_j^*<1$ if $U_i(\be_j)>\tau_i$, and $\lambda_j^*=1$ if $U_i(\be_j)=\tau_i$. Hence, with $\mathcal{O}(m\log(1/\varepsilon))$ queries, we can tell which of the following three cases applies.

\textbf{Case 1: $\lambda_a^*<1$ for some $a\in M_i$.} Then $U_i(\be_a)>\tau_i$. For each $j\in[m]\setminus\{a\}$, let $\mathbf{y}_j$ be the boundary point on the edge from $\be_b$ to $\be_j$ (found above) if $j\in M_i$, and on the edge from $\be_j$ to $\be_a$ if $j\notin M_i$. Each $\mathbf{y}_j$ lies on $H_i$.

We claim that the $m-1$ points $\mathbf{y}_j$ are affinely independent. For each $j\in[m]\setminus\{a,b\}$, the point $\mathbf{y}_j$ is a convex combination of $\be_j$ and one of $\be_a,\be_b$ with a positive coefficient on $\be_j$: if $j\in M_i$, this coefficient is positive because $\be_b$ is rejected and therefore $\mathbf{y}_j\neq\be_b$, and if $j\notin M_i$, it is positive because $\be_a$ is strictly accepted and therefore $\mathbf{y}_j\neq\be_a$. Moreover, $\mathbf{y}_b$ is a combination of $\be_b$ and $\be_a$. Hence, for $j\in[m]\setminus\{a,b\}$, the $j$-th coordinate of $\mathbf{y}_j-\mathbf{y}_b$ is positive, while for every $k\in[m]\setminus\{a,b,j\}$, the $j$-th coordinate of $\mathbf{y}_k-\mathbf{y}_b$ is zero. Thus the $m-2$ vectors $\mathbf{y}_j-\mathbf{y}_b$ with $j\in[m]\setminus\{a,b\}$ are linearly independent, which proves the claim.

Since $H_i$ has dimension $m-2$, it is the affine hull of these points. Hence $\hat u_i$ is the unique vector satisfying
\[
\|\hat u_i\|=1,\quad \sum_{\ell=1}^m\hat u_i(s_\ell)=0,\quad \langle\hat u_i,\mathbf{y}_j-\mathbf{y}_b\rangle=0\ \text{ for all } j\in[m]\setminus\{a,b\},\quad \langle\hat u_i,\be_a-\mathbf{y}_b\rangle>0,
\]
where the equalities determine $\hat u_i$ up to sign, and the last inequality fixes the sign because $\be_a$ is strictly accepted while $\mathbf{y}_b$ lies on the boundary. Finally, $\hat\tau_i=\langle\hat u_i,\mathbf{y}_b\rangle$.

\textbf{Case 2: $\lambda_j^*=1$ for every $j\in M_i$, and $|M_i|=m-1$.} Then $U_i(\be_j)=\tau_i$ for every $j\in M_i$, and $U_i(\be_b)<\tau_i$ for the unique $b\notin M_i$. For any $\bx\in\Delta(S)$,
\[
U_i(\bx)=\sum_{j\in M_i}x_jU_i(\be_j)+x_bU_i(\be_b)=\tau_i(1-x_b)+x_bU_i(\be_b)\le\tau_i,
\]
with equality if and only if $x_b=0$. Therefore, $A_i=\{\bx\in\Delta(S):x_b=0\}=\mathrm{conv}\{\be_j:j\in M_i\}$. The $m-1$ affinely independent points $\be_j$ with $j\in M_i$ lie on $H_i$, so $H_i$ is their affine hull, and $\hat u_i$ is determined up to sign by $\|\hat u_i\|=1$, $\sum_{\ell=1}^m\hat u_i(s_\ell)=0$, and $\langle\hat u_i,\be_j-\be_{j'}\rangle=0$ for all $j,j'\in M_i$. We choose the sign so that $\hat u_i(s_b)<\hat u_i(s_j)$ for $j\in M_i$, and set $\hat\tau_i:=\hat u_i(s_j)$ for any $j\in M_i$.

\textbf{Case 3: $\lambda_j^*=1$ for every $j\in M_i$, and $|M_i|\le m-2$.} Then $i\in B$. As in Case~2, $U_i(\be_j)=\tau_i$ for every $j\in M_i$ and $U_i(\be_k)<\tau_i$ for every $k\notin M_i$, so for any $\bx\in\Delta(S)$,
\[
U_i(\bx)=\tau_i\sum_{j\in M_i}x_j+\sum_{k\notin M_i}x_kU_i(\be_k)\le\tau_i,
\]
with equality if and only if $x_k=0$ for all $k\notin M_i$. Hence
\begin{equation}\label{eq:face}
A_i=\{\bx\in\Delta(S):x_k=0\text{ for all }k\notin M_i\}=\mathrm{conv}\{\be_j:j\in M_i\},
\end{equation}
which is known once $M_i$ is known.

In each case, the procedure makes $m$ queries to pure lotteries and $\mathcal{O}(m)$ binary searches with $\mathcal{O}(\log(1/\varepsilon))$ queries each, for a total of $\mathcal{O}(m\log(1/\varepsilon))$ queries. Running it for all agents gives $\mathcal{O}(nm\log(1/\varepsilon))$ queries.

\textbf{Agents in $B$.} Let $i\in B$. By Lemma~\ref{lem:largest-violation}, the normalized boundaries that give the same answers as agent $i$ correspond to the positive vectors $(\delta_j)_{j\notin M_i}$ with $\sum_{j\notin M_i}\delta_j^2-\frac1m(\sum_{j\notin M_i}\delta_j)^2=1$, and different vectors give different violations. By~\eqref{eq:G-positive} below, the left-hand side of this equation is positive for every nonzero vector, so every positive vector becomes such a vector after scaling. Since agent $i$ rejects at least two pure alternatives, there are infinitely many positive vectors, no two of which are proportional, and they give infinitely many different normalized boundaries.

\textbf{Lower bound.} Let $m\ge 3$, and consider an agent with $u(s_1)=0$, $u(s_2)=1$, $u(s_j)\in\{0,\varepsilon,\dots,1\}$ for $j\in\{3,\dots,m\}$, and $\tau\in\{\varepsilon,2\varepsilon,\dots,1-\varepsilon\}$. Such an agent rejects $\be_1$ and strictly accepts $\be_2$, so she is not in $B$. Different choices of $(u,\tau)$ give different normalized boundaries. Indeed, if $(u,\tau)$ and $(u',\tau')$ have the same normalized boundary, then their centered vectors are positive multiples of each other, which means that $u'=cu+d\mathbf{1}$ and $\tau'=c\tau+d$ for some $c>0$ and $d\in\mathbb{R}$; the conditions $u(s_1)=u'(s_1)=0$ and $u(s_2)=u'(s_2)=1$ then force $d=0$ and $c=1$. Hence there are $(1/\varepsilon-1)(1/\varepsilon+1)^{m-2}\ge(1/\varepsilon)^{m-2}$ such agents with pairwise different normalized boundaries, and at least $(1/\varepsilon)^{n(m-2)}$ profiles of $n$ such agents, all $\varepsilon$-quantized and with $B=\varnothing$. Consider a deterministic algorithm that finds the normalized boundaries of all agents, and suppose that it makes at most $L$ queries on each of these profiles. Since each query has two possible answers, the algorithm observes at most $2^L$ different sequences of answers, and its output is determined by this sequence. As it must give different outputs on different profiles, we have $2^L\ge(1/\varepsilon)^{n(m-2)}$, so $L\ge n(m-2)\log_2(1/\varepsilon)\ge\frac{nm}{3}\log_2(1/\varepsilon)$, where the last inequality uses $m\ge 3$.

\subsection{Proof of Lemma~\ref{lem:largest-violation}}\label{app:proof-largest-violation}

Let $i\in B$ and $C:=[m]\setminus M_i$, so that $|C|\ge 2$. By~\eqref{eq:face}, $A_i=\{\bx\in\Delta(S):x_j=0\text{ for all }j\in C\}$. A normalized boundary gives the same answer as agent $i$ to every membership query if and only if its acceptable set within $\Delta(S)$ is $A_i$. We first describe all such boundaries, and then maximize the violation over them.

\textbf{Consistent boundaries.} Let $(\hat u,\hat\tau)$ be a pair with $\|\hat u\|=1$ and $\sum_{j=1}^m\hat u(s_j)=0$ such that $\{\mathbf{y}\in\Delta(S):\langle\hat u,\mathbf{y}\rangle\ge\hat\tau\}=A_i$. Since $\be_j\in A_i$ for $j\in M_i$, we have $\hat u(s_j)\ge\hat\tau$ for all $j\in M_i$. We claim that, in fact, $\hat u(s_j)=\hat\tau$ for all $j\in M_i$. Indeed, if $\hat u(s_j)>\hat\tau$ for some $j\in M_i$, then for any $k\in C$, we have $(1-\lambda)\hat u(s_j)+\lambda\hat u(s_k)>\hat\tau$ for all sufficiently small $\lambda>0$, so the lottery $(1-\lambda)\be_j+\lambda\be_k\notin A_i$ would be accepted, a contradiction. Also, $\be_j\notin A_i$ for $j\in C$, so $\hat u(s_j)<\hat\tau$. Let $\delta_j:=\hat\tau-\hat u(s_j)>0$ for $j\in C$. Since $\sum_{j=1}^m\hat u(s_j)=0$, we have $m\hat\tau=\sum_{j\in C}\delta_j$, and since $\|\hat u\|=1$,
\begin{equation}\label{eq:delta-constraint}
1=\sum_{j=1}^m\hat u(s_j)^2=\sum_{j=1}^m(\hat\tau-\hat u(s_j))^2-m\hat\tau^2=\sum_{j\in C}\delta_j^2-\frac1m\Big(\sum_{j\in C}\delta_j\Big)^2.
\end{equation}
Moreover, for every $\bx\in\Delta(S)$, using $\sum_{j=1}^m x_j=1$,
\[
\hat\tau-\langle\hat u,\bx\rangle=\sum_{j=1}^m x_j(\hat\tau-\hat u(s_j))=\sum_{j\in C}\delta_jx_j\ge 0,
\]
so the violation of $(\hat u,\hat\tau)$ at $\bx$ is $\sum_{j\in C}\delta_jx_j$.

Conversely, let $\delta\in\mathbb{R}^C$ be a positive vector satisfying~\eqref{eq:delta-constraint}, and set $\hat\tau:=\frac1m\sum_{j\in C}\delta_j$, $\hat u(s_j):=\hat\tau$ for $j\in M_i$, and $\hat u(s_j):=\hat\tau-\delta_j$ for $j\in C$. Then $\sum_{j=1}^m\hat u(s_j)=m\hat\tau-\sum_{j\in C}\delta_j=0$, and the calculation in~\eqref{eq:delta-constraint} gives $\|\hat u\|=1$. As above, the violation at $\bx\in\Delta(S)$ is $\sum_{j\in C}\delta_jx_j$, which is zero if and only if $x_j=0$ for all $j\in C$, that is, if and only if $\bx\in A_i$. This proves the first part of the lemma. By Appendix~\ref{app:halfspace-assumption}, each of these normalized boundaries is induced by some utility vector in $[0,1]^m$ and threshold in $(0,1]$.

\textbf{The largest violation.} Fix $\bx\in\Delta(S)$ and let $\mathbf{z}:=(x_j)_{j\in C}\in\mathbb{R}^C_{\ge 0}$. By the first part, we need to maximize $\langle\mathbf{z},\delta\rangle$ over positive vectors $\delta\in\mathbb{R}^C$ with $\delta^\top G\delta=1$, where $G:=I-\frac1m\mathbf{1}\mathbf{1}^\top$ is a $|C|\times|C|$ matrix and $\mathbf{1}$ is the all-ones vector in $\mathbb{R}^C$. For every nonzero $\delta\in\mathbb{R}^C$, the Cauchy--Schwarz inequality gives $(\mathbf{1}^\top\delta)^2\le|C|\,\|\delta\|^2$, so
\begin{equation}\label{eq:G-positive}
\delta^\top G\delta=\|\delta\|^2-\frac1m(\mathbf{1}^\top\delta)^2\ge\Big(1-\frac{|C|}{m}\Big)\|\delta\|^2=\frac{|M_i|}{m}\|\delta\|^2>0.
\end{equation}
Thus $G$ is positive definite, and it is easy to verify that $G^{-1}=I+\frac{1}{|M_i|}\mathbf{1}\mathbf{1}^\top$. By the Cauchy--Schwarz inequality for the inner product induced by $G$,
\[
\langle\mathbf{z},\delta\rangle=(G^{-1}\mathbf{z})^\top G\delta\le\sqrt{\mathbf{z}^\top G^{-1}\mathbf{z}}\,\sqrt{\delta^\top G\delta}=\sqrt{\mathbf{z}^\top G^{-1}\mathbf{z}}.
\]
If $\mathbf{z}=0$, then every consistent boundary has violation $0$ at $\bx$. Otherwise, equality holds for $\delta=G^{-1}\mathbf{z}/\sqrt{\mathbf{z}^\top G^{-1}\mathbf{z}}$, which is positive because $G^{-1}\mathbf{z}=\mathbf{z}+\frac{1}{|M_i|}(\mathbf{1}^\top\mathbf{z})\mathbf{1}$ and $\mathbf{1}^\top\mathbf{z}>0$. In both cases, the largest violation at $\bx$ is
\begin{equation}\label{eq:largest-violation}
\sqrt{\mathbf{z}^\top G^{-1}\mathbf{z}}=\sqrt{\sum_{j\in C}x_j^2+\frac{1}{|M_i|}\Big(\sum_{j\in C}x_j\Big)^2}.
\end{equation}

On the other hand, define $\mathbf{y}\in\mathbb{R}^m$ by $y_j:=0$ for $j\in C$ and $y_j:=x_j+\frac{1}{|M_i|}\sum_{\ell\in C}x_\ell$ for $j\in M_i$. Then $\mathbf{y}\ge 0$, $\sum_{j=1}^m y_j=1$, and $y_j=0$ for $j\in C$, so $\mathbf{y}\in A_i$. Moreover, $\bx-\mathbf{y}$ is orthogonal to every vector $\mathbf{h}\in\mathbb{R}^m$ with $h_j=0$ for $j\in C$ and $\sum_{j\in M_i}h_j=0$, since $\bx-\mathbf{y}$ is constant on the coordinates in $M_i$. These vectors $\mathbf{h}$ are exactly the directions of the affine hull of $A_i$, so $\mathbf{y}$ is the orthogonal projection of $\bx$ onto this affine hull. Since $\mathbf{y}\in A_i$, it is also the closest point of $A_i$ to $\bx$, and therefore
\[
\mathrm{dist}(\bx,A_i)^2=\|\bx-\mathbf{y}\|^2=\sum_{j\in C}x_j^2+\frac{1}{|M_i|}\Big(\sum_{j\in C}x_j\Big)^2.
\]
Comparing with~\eqref{eq:largest-violation} shows that the largest violation at $\bx$ equals $\mathrm{dist}(\bx,A_i)$.

Finally, $\bx\mapsto\mathrm{dist}(\bx,A_i)^2$ is convex, so its maximum over $\Delta(S)$ is attained at a pure lottery. The formula above gives the value $1+1/|M_i|$ at $\be_j$ for $j\in C$, and $0$ at $\be_j$ for $j\in M_i$. Hence $\max_{\bx\in\Delta(S)}\mathrm{dist}(\bx,A_i)=\sqrt{1+1/|M_i|}$.

\subsection{Proof of Theorem~\ref{thm:worst-case-objective}}\label{app:proof-worst-case}

We first justify the two claims about $\bar F$ made before the theorem. For each $i\notin B$, the function $v_i$ is convex, and since $t\mapsto t^p$ is convex and nondecreasing on $\mathbb{R}_{\ge 0}$, so is $v_i^p$. For each $i\in B$, the function $\bx\mapsto\mathrm{dist}(\bx,A_i)$ is convex because $A_i$ is a nonempty closed convex set, and so is its $p$-th power. Hence $\bar F$ is convex and continuous on the compact set $\Delta(S)$, so a minimizer $\bar\bx$ exists. Moreover, the normalized boundaries of the agents not in $B$ are determined by the answers (Theorem~\ref{thm:identification}), the boundaries of different agents can be chosen independently of each other, and, for each $i\in B$, the largest violation at $\bx$ consistent with the answers is $\mathrm{dist}(\bx,A_i)$ (Lemma~\ref{lem:largest-violation}). Therefore, $\bar F(\bx)$ is the largest value of $V_p(\bx)^p$ over all profiles that give the same answers as the true profile.

In particular, $V_p(\bx)^p\le\bar F(\bx)$ for every $\bx\in\Delta(S)$, with $\bar F(\bx)-V_p(\bx)^p=\frac1n\sum_{i\in B}(\mathrm{dist}(\bx,A_i)^p-v_i(\bx)^p)$. A minimizer $\bx^*$ of $V_p$ exists since $V_p$ is continuous on $\Delta(S)$. Hence
\[
V_p(\bar\bx)^p\le\bar F(\bar\bx)\le\bar F(\bx^*)=V_p(\bx^*)^p+\frac1n\sum_{i\in B}\left(\mathrm{dist}(\bx^*,A_i)^p-v_i(\bx^*)^p\right),
\]
which is the first bound. By Lemma~\ref{lem:largest-violation}, $\mathrm{dist}(\bx^*,A_i)\le\sqrt{1+1/|M_i|}\le\sqrt{2}$ for every $i\in B$, so
\[
V_p(\bar\bx)^p\le V_p(\bx^*)^p+\frac{|B|}{n}(\sqrt{2})^p.
\]
Taking $p$-th roots and using $(a+b)^{1/p}\le a^{1/p}+b^{1/p}$ for $a,b\ge 0$ gives $V_p(\bar\bx)\le V_p(\bx^*)+\sqrt{2}(|B|/n)^{1/p}$. Replacing $\sqrt{2}$ by $\max_{i\in B}\sqrt{1+1/|M_i|}$ in the last two steps gives the refinement mentioned after the theorem.

\subsection{Proof of Theorem~\ref{thm:ambiguity-lower-bound}}\label{app:proof-ambiguity-lb}

The high-level idea is to construct two profiles that give the same answer to every query and are mirror images of each other under swapping $s_1$ and $s_2$. Each profile has a good lottery that leans toward one of these two alternatives, while, by convexity, every lottery is on average at least as bad as a symmetric one, which is far from optimal in both profiles. We first consider two agents and $m=3$, and then add more agents. All thresholds are $1$. In the first profile, the utility vectors of agents~$1$ and~$2$ are $(1,0,\tfrac12)$ and $(\tfrac12,1,0)$; in the second profile, they are $(1,\tfrac12,0)$ and $(0,1,\tfrac12)$. In both profiles, agent~$1$ accepts only $\be_1$ and agent~$2$ accepts only $\be_2$, so the two profiles give the same answer to every membership query, and both agents belong to $B$. Centering and normalizing as in Section~\ref{sec:preliminaries} gives the following violations at $\bx\in\Delta(S)$:
\begin{align*}
\text{first profile:}&\quad v_1(\bx)=\sqrt{2}\,x_2+\tfrac{1}{\sqrt{2}}\,x_3,\quad v_2(\bx)=\tfrac{1}{\sqrt{2}}\,x_1+\sqrt{2}\,x_3;\\
\text{second profile:}&\quad v_1(\bx)=\tfrac{1}{\sqrt{2}}\,x_2+\sqrt{2}\,x_3,\quad v_2(\bx)=\sqrt{2}\,x_1+\tfrac{1}{\sqrt{2}}\,x_3.
\end{align*}
For instance, in the first profile, agent~$1$ has shortfall $1-x_1-\tfrac12x_3=x_2+\tfrac12x_3$ and centered vector $(\tfrac12,-\tfrac12,0)$ of norm $1/\sqrt{2}$. Let $W_1(\bx)$ and $W_2(\bx)$ be the value of $V_p(\bx)$ in the first and second profile, respectively.

\textbf{The optimal values are small.} In the first profile, both violations at $(\tfrac23,\tfrac13,0)$ equal $\sqrt{2}/3$, so $W_1(\tfrac23,\tfrac13,0)=\sqrt{2}/3$. Similarly, $W_2(\tfrac13,\tfrac23,0)=\sqrt{2}/3$. Hence the minimum of each of $W_1$ and $W_2$ over $\Delta(S)$ is at most $\sqrt{2}/3$.

\textbf{Every lottery is far from optimal in one profile.} Let $\sigma(\bx):=(x_2,x_1,x_3)$. The violations of the first profile at $\sigma(\bx)$ are those of the second profile at $\bx$ in the opposite order, so $W_1(\sigma(\bx))=W_2(\bx)$ and, similarly, $W_2(\sigma(\bx))=W_1(\bx)$. Hence $W_1+W_2$ is invariant under $\sigma$. Since $W_1+W_2$ is also convex, for every $\bx\in\Delta(S)$,
\[
W_1(\bx)+W_2(\bx)=\frac12\Big((W_1+W_2)(\bx)+(W_1+W_2)(\sigma(\bx))\Big)\ge W_1(\mathbf{y})+W_2(\mathbf{y}),
\]
where $\mathbf{y}:=\frac12(\bx+\sigma(\bx))=\big(\frac{1-s}{2},\frac{1-s}{2},s\big)$ with $s:=x_3$.
At $\mathbf{y}$, the two violations in each profile are $\frac{1}{\sqrt{2}}$ and $\frac{1+3s}{2\sqrt{2}}\ge\frac{1}{2\sqrt{2}}$. Since the power mean is nondecreasing in each argument and at least the arithmetic mean, $W_1(\mathbf{y})=W_2(\mathbf{y})\ge\frac12\big(\frac{1}{\sqrt{2}}+\frac{1}{2\sqrt{2}}\big)=\frac{3}{4\sqrt{2}}$. Therefore,
\[
W_1(\bx)+W_2(\bx)\ge\frac{3}{2\sqrt{2}}\quad\text{for every }\bx\in\Delta(S).
\]

\textbf{More agents.} Let $b$ be even with $2\le b\le n$. The first (resp., second) profile with $n$ agents consists of $b/2$ copies of the two agents of the first (resp., second) profile above, together with $n-b$ agents with utility vector $(1,1,0)$ and threshold $1$. Each of the added agents accepts exactly the lotteries with $x_3=0$; she accepts $\be_1$ and $\be_2$ on her boundary and rejects only $\be_3$, so she is not in $B$. Hence $|B|=b$ in both profiles, and the two profiles give the same answer to every query. Let $V^{(1)}$ and $V^{(2)}$ denote the objective $V_p$ in the two profiles. Since the added agents have zero violation when $x_3=0$, we have
\[
\min_{\bx\in\Delta(S)}V^{(1)}(\bx)\le V^{(1)}(\tfrac23,\tfrac13,0)=\Big(\frac bn\Big)^{1/p}W_1(\tfrac23,\tfrac13,0)=\Big(\frac bn\Big)^{1/p}\frac{\sqrt{2}}{3},
\]
and similarly for $V^{(2)}$. Moreover, $V^{(1)}(\bx)\ge(b/n)^{1/p}W_1(\bx)$ and $V^{(2)}(\bx)\ge(b/n)^{1/p}W_2(\bx)$ for every $\bx$, since the added agents contribute nonnegative terms. Combining these bounds, the additive losses of any lottery $\bx$ in the two profiles sum to at least
\[
\Big(\frac bn\Big)^{1/p}\left(W_1(\bx)+W_2(\bx)-\frac{2\sqrt{2}}{3}\right)\ge\Big(\frac bn\Big)^{1/p}\left(\frac{3}{2\sqrt{2}}-\frac{2\sqrt{2}}{3}\right)=\frac{1}{6\sqrt{2}}\Big(\frac bn\Big)^{1/p}.
\]

Finally, an algorithm that only uses membership queries receives the same answers in the two profiles. Hence, for every realization of its internal randomness, it returns the same lottery in both profiles, and the distribution of its output $\hat\bx$ is the same. Taking expectations over $\hat\bx$, the expected additive losses in the two profiles sum to at least $\frac{1}{6\sqrt{2}}(b/n)^{1/p}$, so one of them is at least $\frac{1}{12\sqrt{2}}(b/n)^{1/p}$.

We chose the constant $\frac{1}{12\sqrt{2}}$ for simplicity, as it holds for all $p$. For specific values of $p$, the same construction gives larger constants: a short calculation shows that, for $b=n=2$, every lottery has additive loss at least $\frac{1}{4\sqrt{2}}$ for $p=1$ and at least $\frac{1}{4\sqrt{5}}$ for $p=2$ in one of the two profiles.

\subsection{Proof of Corollary~\ref{cor:exact}}\label{app:proof-exact}

Run the algorithm of Theorem~\ref{thm:identification}. Since $B=\varnothing$, it finds all normalized boundaries with $\mathcal{O}(nm\log(1/\varepsilon))$ queries, so every $v_i$ is known on $\Delta(S)$. It remains to minimize $V_p$, or equivalently $F$, over $\Delta(S)$. This is equivalent to the convex program
\[
\begin{array}{ll}
\text{minimize} & \frac1n\sum_{i\in N} t_i^p\\[1mm]
\text{subject to} & t_i \ge \hat\tau_i-\langle \hat u_i,\bx\rangle \ \text{ and } \ t_i\ge 0 \quad \text{for all } i\in N,\\
& \bx\in\Delta(S).
\end{array}
\tag{$\mathrm{CP}_p$}
\]
Indeed, for any $\bx\in\Delta(S)$, choosing $t_i:=v_i(\bx)$ gives a feasible solution with objective value $F(\bx)$; conversely, every feasible $(\bx,t)$ satisfies $t_i\ge v_i(\bx)$ for all $i$, so its objective value is at least $F(\bx)$. Thus the optimal value of $(\mathrm{CP}_p)$ is $\min_{\bx\in\Delta(S)}F(\bx)$, which is attained because $F$ is continuous on the compact set $\Delta(S)$, and for every optimal solution $(\hat\bx,\hat t)$ of $(\mathrm{CP}_p)$, the lottery $\hat\bx$ minimizes $V_p$. The algorithm outputs such a lottery $\hat\bx$.

\subsection{Duality for $1<p<\infty$ and Proof of Theorem~\ref{thm:dual-finite-p}}\label{app:dual}

Theorem~\ref{thm:dual-finite-p} follows from the following duality result for the convex program $(\mathrm{CP}_p)$, in which the factor $1/n$ is absorbed into the dual variables.

\begin{theorem}\label{thm:dual-full}
    Fix $p\in(1,\infty)$ and let $q:=p/(p-1)$. Then
    \[
    \min_{\bx\in\Delta(S)} F(\bx)
    = \max_{\lambda\in\mathbb{R}^n_{\ge 0}}
    \frac1n\left(
    \sum_{i\in N}\left(\lambda_i\hat\tau_i-\frac{p-1}{p^q}\lambda_i^q\right)
    -
    \max_{j\in[m]}\sum_{i\in N}\lambda_i\hat u_i(s_j)
    \right).
    \]
    Moreover, $\bx^*\in\Delta(S)$ minimizes $F$ over $\Delta(S)$ and
    $\lambda^*\in\mathbb{R}^n_{\ge 0}$ is optimal for the problem on the right-hand side if and only if $\lambda_i^* = p\,v_i(\bx^*)^{p-1}$ for all $i \in N$
    and every $j\in[m]$ with $x_j^*>0$ satisfies $\sum_{i\in N}\lambda_i^*\hat u_i(s_j)
    =
    \max_{\ell\in[m]}\sum_{i\in N}\lambda_i^*\hat u_i(s_\ell)$.
\end{theorem}

\begin{proof}
    We work with the Lagrangian
    \[
    L(\bx,t,\lambda)=\frac1n\sum_{i\in N}\left(t_i^p+\lambda_i(\hat\tau_i-\langle \hat u_i,\bx\rangle-t_i)\right),
    \]
    where $\bx\in\Delta(S)$, $t\in\mathbb R_{\ge 0}^n$, and $\lambda\in\mathbb{R}^n_{\ge 0}$. Since $(\mathrm{CP}_p)$ is a convex program and Slater's condition holds (choose any $\bx$ in the relative interior of $\Delta(S)$ and each $t_i$ larger than $\max\{0,\hat{\tau}_i-\langle \hat{u}_i,\bx\rangle\}$), strong duality holds, and the dual optimum is attained.

    For fixed $\lambda$, minimizing over $\bx$ and $t$ gives
    \begin{equation*}
        \inf_{\bx \in \Delta(S),\, t \in \mathbb{R}_{\ge 0}^n} L(\bx,t,\lambda)
        = \frac{1}{n}\sum_{i \in N} \lambda_i \hat{\tau}_i
        - \max_{j \in [m]} \frac{1}{n}\sum_{i \in N}\lambda_i \hat{u}_i(s_j)
        + \frac{1}{n}\sum_{i \in N}\inf_{r \ge 0} (r^p - \lambda_i r),
    \end{equation*}
    because a linear function on $\Delta(S)$ attains its minimum at a pure lottery.
    For $\lambda_i \ge 0$, the function $r \mapsto r^p - \lambda_i r$
    is strictly convex on $\mathbb{R}_{\ge 0}$, its minimizer is $(\lambda_i / p)^{1/(p-1)}$, and its minimum value is $-\frac{p-1}{p^q}\lambda_i^q$. This gives the dual problem in the statement.

    Now let $\bx^*$ minimize $F$, let $t_i^* := v_i(\bx^*)$ for all $i\in N$, and let $\lambda^*$ be dual-optimal. By the proof of Corollary~\ref{cor:exact}, $(\bx^*,t^*)$ is optimal for $(\mathrm{CP}_p)$. By strong duality, $(\bx^*,t^*,\lambda^*)$ is a saddle point of $L$. Hence each $t_i^*$ minimizes $r\mapsto r^p-\lambda_i^* r$ over $\mathbb{R}_{\ge 0}$, so $t_i^*=(\lambda_i^*/p)^{1/(p-1)}$, that is, $\lambda_i^* = p\,v_i(\bx^*)^{p-1}$. Also, $\bx^*$ maximizes $\bx \mapsto \sum_{i \in N}\lambda_i^* \langle \hat{u}_i, \bx\rangle$ over $\Delta(S)$. A linear function on the simplex is maximized exactly by the lotteries that only use alternatives with the largest coefficient, so every $j \in [m]$ with $x_j^*>0$ satisfies $\sum_{i \in N}\lambda_i^* \hat{u}_i(s_j) = \max_{\ell \in [m]} \sum_{i \in N}\lambda_i^* \hat{u}_i(s_\ell)$.

    Conversely, suppose that $\bx^* \in \Delta(S)$ and $\lambda^* \in \mathbb{R}_{\ge 0}^n$ satisfy the two conditions, and let $t_i^* := v_i(\bx^*)$. Then $(\bx^*,t^*)$ is feasible for $(\mathrm{CP}_p)$. For each $i \in N$, since $\lambda_i^* = p\,(t_i^*)^{p-1}$, we have $(t_i^*)^p = \lambda_i^* t_i^* - \frac{p-1}{p^q}(\lambda_i^*)^q$. Moreover, $\lambda_i^*(\hat{\tau}_i - \langle \hat{u}_i, \bx^*\rangle) = \lambda_i^* t_i^*$: if $t_i^*>0$, then $t_i^* = \hat{\tau}_i - \langle \hat{u}_i, \bx^*\rangle$, and if $t_i^* = 0$, then $\lambda_i^* = 0$. Finally, because every $j$ with $x_j^*>0$ attains the largest score, $\sum_{i\in N}\lambda_i^*\langle \hat u_i, \bx^*\rangle = \max_{\ell\in[m]}\sum_{i\in N}\lambda_i^*\hat u_i(s_\ell)$. Combining these equalities, we obtain
    \begin{equation*}
        F(\bx^*)
        = \frac{1}{n}\sum_{i \in N}(t_i^*)^p
        = \frac{1}{n}\sum_{i \in N}
        \left(
        \lambda_i^* \hat{\tau}_i - \frac{p-1}{p^q}(\lambda_i^*)^q
        \right)
        - \max_{j \in [m]} \frac{1}{n}\sum_{i \in N}\lambda_i^* \hat{u}_i(s_j),
    \end{equation*}
    that is, the primal objective value at $(\bx^*,t^*)$ equals the dual objective value at $\lambda^*$. By weak duality, $\bx^*$ minimizes $F$ and $\lambda^*$ is dual-optimal.
\end{proof}

\begin{proof}[Proof of Theorem~\ref{thm:dual-finite-p}]
    Minimizing $V_p$ is equivalent to minimizing $F$. If $\bx^*$ minimizes $F$, let $\lambda^*$ be dual-optimal; by Theorem~\ref{thm:dual-full}, $\lambda_i^*=p\,v_i(\bx^*)^{p-1}$, and every alternative that $\bx^*$ uses maximizes $\sum_{i\in N}\lambda_i^*\hat u_i(s_j)=p\sum_{i\in N}v_i(\bx^*)^{p-1}\hat u_i(s_j)$. Conversely, if every alternative that $\bx^*$ uses maximizes $\sum_{i\in N}v_i(\bx^*)^{p-1}\hat u_i(s_j)$, then $\bx^*$ and $\lambda_i^*:=p\,v_i(\bx^*)^{p-1}$ satisfy both conditions of Theorem~\ref{thm:dual-full}, so $\bx^*$ minimizes $F$. Finally, fix any minimizer $\bx^*$; by Theorem~\ref{thm:dual-full}, every dual-optimal solution equals $(p\,v_i(\bx^*)^{p-1})_{i\in N}$, so the dual optimum is unique.
\end{proof}

\subsection{The Cases $p=1$ and $p=\infty$}\label{app:endpoint-kkt}

This subsection shows that the description after Theorem~\ref{thm:dual-finite-p} also holds for $p=1$ and $p=\infty$, with different rules for the weights. In both cases, every optimal lottery only uses alternatives that maximize a weighted score, where the weights form an optimal solution of a dual problem. For $p=1$, the weights lie in $[0,1]$, and every violated agent has weight $1$. For $p=\infty$, the weights sum to at most $1$, and if the optimal value is positive, only the most violated agents can have positive weight.

\begin{theorem}[Duality for $p=1$]\label{thm:p1-dual}
	Let $p=1$, so that $F=V_1$. Then $\min_{\bx\in\Delta(S)} F(\bx) = \max_{\lambda\in[0,1]^n} \left( \frac1n\sum_{i\in N}\lambda_i \hat\tau_i - \max_{j\in[m]}\frac1n\sum_{i\in N}\lambda_i \hat u_i(s_j) \right)$.

	Moreover, let $\bx^*\in\Delta(S)$ and $\lambda^*\in[0,1]^n$, and set $t_i^*:=v_i(\bx^*)$ for all $i \in N$.
	Then $\bx^*$ minimizes $F$ over $\Delta(S)$ and $\lambda^*$ is dual-optimal if and only if 
	\begin{enumerate}[label=(\roman*)]
		\item $\lambda_i^*(\hat\tau_i-\langle \hat u_i,\bx^*\rangle-t_i^*)=0$ for all $i \in N$;
		\item $(1-\lambda_i^*)\,t_i^*=0$ for all $i \in N$; and 
		\item every $j\in[m]$ with $x_j^*>0$ satisfies $\sum_{i\in N}\lambda_i^* \hat u_i(s_j) = \max_{\ell\in[m]}\sum_{i\in N}\lambda_i^* \hat u_i(s_\ell)$.
	\end{enumerate}

In particular,
\[
v_i(\bx^*)>0 \implies \lambda_i^*=1,
\quad
\hat\tau_i-\langle \hat u_i,\bx^*\rangle<0 \implies \lambda_i^*=0.
\]
In words, every violated agent has weight $1$, every agent who strictly accepts $\bx^*$ has weight $0$, and every agent whose boundary contains $\bx^*$ may have any weight in $[0,1]$ that is consistent with conditions~(i)--(iii).
\end{theorem}

\begin{proof}
    Consider the epigraph reformulation
    \[
    \begin{array}{ll}
    \text{minimize} & \frac1n \sum_{i\in N} t_i \\[1mm]
    \text{subject to} & t_i \ge \hat\tau_i - \langle \hat u_i, \bx\rangle \quad \text{for each } i \in N,\\
    & t_i \ge 0 \quad \text{for each } i \in N,\\
    & \bx\in\Delta(S).
    \end{array}
    \]
    Its optimal value is $\min_{\bx\in\Delta(S)}F(\bx)$, because for fixed $\bx$ the smallest feasible choice is $t_i=v_i(\bx)$.
    
    For $\lambda\in\mathbb{R}^n_{\ge 0}$, consider the Lagrangian
    \[
    L(\bx,t,\lambda)
    =
    \frac1n\sum_{i\in N}(t_i+\lambda_i(\hat\tau_i-\langle \hat u_i,\bx\rangle-t_i)),
    \]
    with domain $\bx\in\Delta(S)$ and $t\in\mathbb{R}^n_{\ge 0}$. 
    As in the proof of Theorem~\ref{thm:dual-full}, Slater's condition holds, so strong duality applies.    
    For fixed $\lambda$, we minimize over $\bx$ as in the proof of Theorem~\ref{thm:dual-full}, and minimizing over each $t_i\ge 0$ gives
    \[
    \inf_{r\ge 0}(1-\lambda_i)r
    =
    \begin{cases}
    0, & \lambda_i\in[0,1],\\
    -\infty, & \lambda_i>1.
    \end{cases}
    \]
    Therefore,
    \[
    \min_{\bx\in\Delta(S)} F(\bx)
    =
    \max_{\lambda\in[0,1]^n}
    \left(
    \frac1n\sum_{i\in N}\lambda_i \hat\tau_i
    -
    \max_{j\in[m]}\frac1n\sum_{i\in N}\lambda_i \hat u_i(s_j)
    \right).
    \]
    The dual feasible set $[0,1]^n$ is compact, so a dual optimum is attained.
    
    Now let $\bx^*$ be primal-optimal and $\lambda^*$ dual-optimal, and set $t_i^*:=v_i(\bx^*)$. 
    By strong duality, $(\bx^*,t^*,\lambda^*)$ is a saddle point. 
    Thus, each $t_i^*$ minimizes $r\mapsto (1-\lambda_i^*)r$ over $r\ge 0$, which is equivalent to
    \[
    (1-\lambda_i^*)\,t_i^*=0
    \quad \text{for each } i\in N.
    \]
    Complementary slackness for the constraints
    $t_i\ge \hat\tau_i-\langle \hat u_i, \bx\rangle$
    gives
    \[
    \lambda_i^*(\hat\tau_i-\langle \hat u_i,\bx^*\rangle-t_i^*)=0
    \quad \text{for each } i\in N.
    \]
    Finally, $\bx^*$ minimizes
    $\bx\mapsto -\sum_{i\in N}\lambda_i^*\langle \hat u_i, \bx\rangle$ over $\Delta(S)$, or equivalently maximizes $\bx\mapsto \sum_{i\in N}\lambda_i^*\langle \hat u_i,\bx\rangle$.
    A linear function on the simplex is maximized exactly by the lotteries that only use alternatives with the largest coefficient, so every $j$ with $x_j^*>0$ satisfies
    \[
    \sum_{i\in N}\lambda_i^* \hat u_i(s_j)
    =
    \max_{\ell\in[m]}\sum_{i\in N}\lambda_i^* \hat u_i(s_\ell).
    \]
    
    Conversely, suppose $\bx^*\in\Delta(S)$ and $\lambda^*\in[0,1]^n$ satisfy the displayed conditions, and set $t_i^*:=v_i(\bx^*)$. Then $(\bx^*,t^*)$ is primal-feasible and $\lambda^*$ is dual-feasible. 
    Since $(1-\lambda_i^*)t_i^*=0$, we have $t_i^*=\lambda_i^* t_i^*$ for every $i$. 
    Using also $\lambda_i^*(\hat\tau_i-\langle \hat u_i, \bx^*\rangle-t_i^*)=0$, we obtain
    \[
    t_i^*=\lambda_i^*t_i^*=\lambda_i^*(\hat\tau_i-\langle \hat u_i, \bx^*\rangle)
    \quad \text{for each } i\in N.
    \]
    Summing over $i$ and using condition~(iii) gives
    \[
    \frac1n\sum_{i\in N} t_i^*
    = \frac1n\sum_{i\in N}\lambda_i^* \hat\tau_i
    - \max_{j\in[m]}\frac1n\sum_{i\in N}\lambda_i^* \hat u_i(s_j).
    \]
    Thus, the primal objective value at $(\bx^*,t^*)$ equals the dual objective value at $\lambda^*$. By weak duality, both are optimal.
    
    The final implications are immediate: if $v_i(\bx^*)=t_i^*>0$, then $(1-\lambda_i^*)t_i^*=0$ forces $\lambda_i^*=1$; and if $\hat\tau_i-\langle \hat u_i, \bx^*\rangle<0$, then complementary slackness forces $\lambda_i^*=0$.
\end{proof}

\begin{theorem}[Duality for $p=\infty$]\label{thm:pinf-dual}
    We have
    \[
    \min_{\bx\in\Delta(S)} V_\infty(\bx)
    = \max_{\lambda\in \mathbb{R}^n_{\ge 0}:\; \sum_{i\in N}\lambda_i \le 1}
    \left(
    \sum_{i\in N}\lambda_i \hat\tau_i
    -
    \max_{j\in[m]} \sum_{i\in N}\lambda_i \hat u_i(s_j)
    \right).
    \]
    Moreover, let $\bx^* \in \Delta(S)$ and $\lambda^* \in \mathbb{R}^n_{\ge 0}$, and set $t^* := V_\infty(\bx^*)$. Then $\bx^*$ minimizes $V_\infty$ over $\Delta(S)$ and $\lambda^*$ is optimal for the displayed dual problem if and only if
    \begin{enumerate}[label=(\roman*)]
        \item $\sum_{i\in N}\lambda_i^* \le 1$;
        \item $\lambda_i^*(\hat{\tau}_i - \langle \hat{u}_i, \bx^* \rangle - t^*) = 0$ for all $i \in N$;
        \item $t^* \left(1 - \sum_{i\in N}\lambda_i^* \right) = 0$; and
        \item  every $j \in [m]$ with $x_j^* > 0$ satisfies $\sum_{i\in N}\lambda_i^* \hat{u}_i(s_j) = \max_{\ell \in [m]} \sum_{i\in N}\lambda_i^* \hat{u}_i(s_\ell)$.
    \end{enumerate}
In particular, if $V_\infty(\bx^*) > 0$, then $\sum_{i\in N}\lambda_i^* = 1$ and
\[
\lambda_i^* > 0 \implies v_i(\bx^*) = V_\infty(\bx^*)
\quad \text{for every } i \in N.
\]
\end{theorem}

\begin{proof}
    Consider the linear program
    \[
    \begin{array}{ll}
    \text{minimize}   & t \\
    \text{subject to} & t \ge \hat\tau_i-\langle \hat u_i, \bx\rangle \quad (i\in N),\\
                      & t \ge 0,\\
                      & \bx \in \Delta(S).
    \end{array}
    \tag{$\mathrm{CP}_\infty$}
    \]
    For any fixed $\bx\in\Delta(S)$, the smallest $t$ for which $(\bx,t)$ is feasible is
    \[
    \max\{0,\max_{i\in N}(\hat\tau_i-\langle \hat u_i,\bx\rangle)\}
    =
    \max_{i\in N}\max\{0,\hat\tau_i-\langle \hat u_i,\bx\rangle\}
    =
    V_\infty(\bx).
    \]
    Thus, the optimal value of $(\mathrm{CP}_\infty)$ is exactly $\min_{\bx\in\Delta(S)}V_\infty(\bx)$, and for every optimal solution $(\bx^*,t^*)$ of $(\mathrm{CP}_\infty)$ we have $t^*=V_\infty(\bx^*)$.
    
    For $\lambda\in \mathbb{R}^n_{\ge 0}$, the Lagrangian of $(\mathrm{CP}_\infty)$ is
    \[
    L(\bx,t,\lambda)
    = t+\sum_{i\in N}\lambda_i(\hat\tau_i-\langle \hat u_i,\bx\rangle-t),
    \]
    with domain $\bx\in\Delta(S)$ and $t\in\mathbb{R}_{\ge 0}$. Equivalently,
    \[
    L(\bx,t,\lambda)
    = \sum_{i\in N}\lambda_i \hat\tau_i
    - \langle \sum_{i\in N}\lambda_i \hat u_i,\bx\rangle
    + t\left(1-\sum_{i\in N}\lambda_i\right).
    \]
    Minimizing over $\bx\in\Delta(S)$ gives
    \[
    \inf_{\bx\in\Delta(S)} L(\bx,t,\lambda)
    = \sum_{i\in N}\lambda_i \hat\tau_i
    - \max_{j\in[m]} \sum_{i\in N}\lambda_i \hat u_i(s_j)
    + t\left(1-\sum_{i\in N}\lambda_i\right),
    \]
    because a linear function on the simplex attains its minimum at a vertex. Therefore
    \[
    \inf_{\bx\in\Delta(S),\,t\in\mathbb{R}_{\ge 0}} L(\bx,t,\lambda)
    =
    \begin{cases}
    \displaystyle
    \sum_{i\in N}\lambda_i \hat\tau_i
    -
    \max_{j\in[m]} \sum_{i\in N}\lambda_i \hat u_i(s_j),
    & \text{if } \sum_{i\in N}\lambda_i \le 1,\\[1.2ex]
    -\infty, & \text{if } \sum_{i\in N}\lambda_i > 1.
    \end{cases}
    \]
    Thus the dual problem is exactly
    \[
    \max_{\lambda\in \mathbb{R}^n_{\ge 0}:\; \sum_{i\in N}\lambda_i \le 1}
    \left(
    \sum_{i\in N}\lambda_i \hat\tau_i
    -
    \max_{j\in[m]} \sum_{i\in N}\lambda_i \hat u_i(s_j)
    \right).
    \]
    
    The primal optimum is attained because $V_\infty$ is continuous on the compact set $\Delta(S)$. The dual feasible set $\{ \lambda\in\mathbb{R}^n_{\ge 0} : \sum_{i\in N}\lambda_i \le 1 \}$ is compact, and the dual objective is continuous, so the dual optimum is attained. 
    Since $(\mathrm{CP}_\infty)$ is a convex program and Slater's condition holds (for example, take $\bx=(1/m,\dots,1/m)$ and choose $t$ strictly larger than $\max\{0,\max_i(\hat\tau_i-\langle \hat u_i,\bx\rangle)\}$), strong duality applies.
    
    We now prove the KKT characterization. 
    Let $(\bx^*,t^*)$ be primal-optimal for $(\mathrm{CP}_\infty)$ and let $\lambda^*$ be dual-optimal. 
    By strong duality, $(\bx^*,t^*,\lambda^*)$ is a saddle point of the Lagrangian. 
    Thus, complementary slackness gives
    \[
    \lambda_i^*(\hat\tau_i-\langle \hat u_i,\bx^*\rangle-t^*)=0
    \quad \text{for each } i\in N.
    \]
    Also, $t^*$ minimizes the function $t \mapsto t(1-\sum_{i\in N}\lambda_i^*)$ over $\mathbb{R}_{\ge 0}$. 
    This is equivalent to
    \[
    \sum_{i\in N}\lambda_i^* \le 1
    \quad\text{and}\quad
    t^*\left(1-\sum_{i\in N}\lambda_i^*\right)=0.
    \]
    Finally, $\bx^*$ minimizes $\bx \mapsto -\sum_{i\in N}\lambda_i^* \langle \hat u_i, \bx\rangle$ over $\Delta(S)$, or equivalently maximizes $\bx \mapsto \sum_{i\in N}\lambda_i^* \langle \hat u_i,\bx\rangle$.
    A linear function on the simplex is maximized exactly by the lotteries that only use alternatives with the largest coefficient, so every $j\in[m]$ with $x_j^*>0$ satisfies
    \[
    \sum_{i\in N}\lambda_i^* \hat u_i(s_j)
    =
    \max_{\ell\in[m]} \sum_{i\in N}\lambda_i^* \hat u_i(s_\ell).
    \]
    
    Conversely, suppose $\bx^*\in\Delta(S)$ and $\lambda^*\in\mathbb{R}^n_{\ge 0}$ satisfy the displayed conditions, and set $t^*:=V_\infty(\bx^*)$. 
    Then $(\bx^*,t^*)$ is primal-feasible for $(\mathrm{CP}_\infty)$, and $\lambda^*$ is dual-feasible. 
    By condition~(iv),
    \[
    \sum_{i\in N}\lambda_i^* \langle \hat u_i, \bx^*\rangle
    = \sum_{j=1}^m x_j^* \sum_{i\in N}\lambda_i^* \hat u_i(s_j)
    = \max_{\ell\in[m]} \sum_{i\in N}\lambda_i^* \hat u_i(s_\ell).
    \]
    Summing condition~(ii) over $i\in N$ and using the previous display, we obtain
    \[
    \sum_{i\in N}\lambda_i^* \hat\tau_i
    -
    \max_{\ell\in[m]} \sum_{i\in N}\lambda_i^* \hat u_i(s_\ell)
    =
    t^* \sum_{i\in N}\lambda_i^*
    =
    t^*,
    \]
    where the last equality follows from condition~(iii).
    Thus, the dual objective value at $\lambda^*$ equals the primal objective value at $(\bx^*,t^*)$. By weak duality, both are optimal.
    
    For the final claim, if $V_\infty(\bx^*)=t^*>0$, then $t^*\left(1-\sum_{i\in N}\lambda_i^*\right)=0$ forces $\sum_{i\in N}\lambda_i^*=1$. And if $\lambda_i^*>0$, then complementary slackness gives $\hat\tau_i-\langle \hat u_i,\bx^*\rangle=t^*>0$ and thus
    \[
    v_i(\bx^*)
    =
    \max\{0,\hat\tau_i-\langle \hat u_i,\bx^*\rangle\}
    =
    t^*
    =
    V_\infty(\bx^*).
    \]
    This proves the theorem.
\end{proof}

\subsection{Proof of Theorem~\ref{thm:few-alternatives}}
    For each alternative $s_j$, let $\mathbf{g}_j := (\hat u_i(s_j))_{i\in N} \in \mathbb{R}^n$, so that $\sum_{j=1}^m x_j\mathbf{g}_j=(\langle \hat u_i, \bx\rangle)_{i\in N}$ for every lottery $\bx$. Since $v_i(\bx)=\max\{0,\hat\tau_i-\langle\hat u_i,\bx\rangle\}$, any two lotteries $\bx$ and $\mathbf{y}$ with $\sum_{j=1}^m x_j\mathbf{g}_j=\sum_{j=1}^m y_j\mathbf{g}_j$ satisfy $v_i(\bx)=v_i(\mathbf{y})$ for every $i\in N$.

    Let $J$ be the set of indices of the alternatives that $\bx^*$ uses. The point $\sum_{j\in J}x_j^*\mathbf{g}_j$ lies in the convex hull of $\{\mathbf{g}_j:j\in J\}\subseteq\mathbb{R}^n$. By Carath\'eodory's theorem, there exist a set $J'\subseteq J$ with $|J'|\le n+1$ and nonnegative numbers $(\mu_j)_{j\in J'}$ with $\sum_{j\in J'}\mu_j=1$ such that $\sum_{j\in J'}\mu_j\mathbf{g}_j=\sum_{j\in J}x_j^*\mathbf{g}_j$. Let $\bar\bx$ be the lottery with $\bar x_j:=\mu_j$ for $j\in J'$ and $\bar x_j:=0$ for $j\notin J'$. Then $\bar\bx$ uses at most $n+1$ alternatives, all of which are used by $\bx^*$, and $v_i(\bar\bx)=v_i(\bx^*)$ for every $i\in N$. Hence $V_p(\bar\bx)=V_p(\bx^*)$, so $\bar\bx$ also minimizes $V_p$.

\subsection{Approximately Optimal Lotteries with Few Alternatives}\label{app:sparse-lotteries}

We continue to assume that the normalized boundaries are known. Theorem~\ref{thm:few-alternatives} shows that some optimal lottery uses at most $n+1$ alternatives. We now give two results in which the number of alternatives does not depend on $n$. First, since $V_p$ is $1$-Lipschitz on $\Delta(S)$, sampling alternatives from an optimal lottery gives a lottery that uses at most $\lceil\alpha^{-2}\rceil$ alternatives and is within additive $\alpha$ of the optimum (Theorem~\ref{thm:random-sparsification}). Second, for $p\ge 2$, the gradient of $F$ is Lipschitz on $\Delta(S)$, so the Frank--Wolfe method reduces $F$ at the usual rate $\mathcal{O}(1/T)$, while its $T$-th iterate uses at most $T+1$ alternatives (Theorem~\ref{thm:fw-sparse}). Each step of this method adds an alternative with the largest weighted score $\sum_{i\in N}v_i(\bx)^{p-1}\hat u_i(s_j)$ at the current lottery $\bx$, which matches the weights in Theorem~\ref{thm:dual-finite-p}.

\begin{theorem}\label{thm:random-sparsification}
    Fix $p\in[1,\infty)$, and let $\bx^*$ be a minimizer of $V_p$ over $\Delta(S)$. For every positive integer $s$, there is a lottery $\bx'$ that uses at most $s$ alternatives, all of which are used by $\bx^*$, such that $V_p(\bx') \le V_p(\bx^*) + 1/\sqrt{s}$. In particular, for every $\alpha>0$, some lottery $\bx'$ that uses at most $\lceil \alpha^{-2}\rceil$ alternatives satisfies $V_p(\bx')\le\min_{\bx\in\Delta(S)}V_p(\bx)+\alpha$.
\end{theorem}

\begin{proof}
    We first show that $V_p$ is $1$-Lipschitz on $\Delta(S)$. Fix $\bx,\mathbf{y}\in\Delta(S)$ and $i\in N$. Since $z\mapsto\max\{0,z\}$ is $1$-Lipschitz and $\|\hat u_i\|=1$, we have $|v_i(\bx)-v_i(\mathbf{y})|\le|\langle\hat u_i,\bx-\mathbf{y}\rangle|\le\|\bx-\mathbf{y}\|$. Hence, by the reverse triangle inequality for the $\ell_p$-norm,
    \[
    |V_p(\bx)-V_p(\mathbf{y})|\le\left(\frac1n\sum_{i\in N}|v_i(\bx)-v_i(\mathbf{y})|^p\right)^{1/p}\le\|\bx-\mathbf{y}\|.
    \]
    Now draw $J_1,\dots,J_s$ independently from the distribution $\bx^*$ on $[m]$, and let $\bar\bx:=\frac1s\sum_{t=1}^s\be_{J_t}$. Then $\bar\bx$ is a lottery that uses at most $s$ alternatives, all of which are used by $\bx^*$. Since $\mathbb{E}[\be_{J_t}]=\bx^*$, the vectors $\be_{J_t}-\bx^*$ are independent with mean zero, so the cross terms vanish and
    \[
    \mathbb{E}\|\bar\bx-\bx^*\|^2=\frac{1}{s^2}\sum_{t=1}^s\mathbb{E}\|\be_{J_t}-\bx^*\|^2=\frac{1-\|\bx^*\|^2}{s}\le\frac1s,
    \]
    where we used $\mathbb{E}\|\be_{J_t}-\bx^*\|^2=\mathbb{E}\|\be_{J_t}\|^2-2\langle\mathbb{E}[\be_{J_t}],\bx^*\rangle+\|\bx^*\|^2=1-\|\bx^*\|^2$. By the Lipschitz bound and Jensen's inequality,
    \[
    \mathbb{E}[V_p(\bar\bx)]\le V_p(\bx^*)+\mathbb{E}\|\bar\bx-\bx^*\|\le V_p(\bx^*)+\sqrt{\mathbb{E}\|\bar\bx-\bx^*\|^2}\le V_p(\bx^*)+\frac{1}{\sqrt{s}}.
    \]
    Therefore some realization $\bx'$ of $\bar\bx$ satisfies $V_p(\bx')\le V_p(\bx^*)+1/\sqrt{s}$. The last sentence of the theorem follows by taking $s=\lceil\alpha^{-2}\rceil$.
\end{proof}

\begin{theorem}\label{thm:fw-sparse}
	Fix $p\in [2,\infty)$ and let $L_p:=p(p-1)2^{(p-2)/2}$. Consider the Frank--Wolfe method for minimizing $F$ over $\Delta(S)$: start from any pure lottery $\bx^{(0)}$, and for each $t\ge 0$, choose $j_t\in \argmax_{j\in [m]} \sum_{i\in N} v_i(\bx^{(t)})^{p-1}\hat u_i(s_j)$ and let $\bx^{(t+1)}:=(1-\gamma_t)\bx^{(t)}+\gamma_t \be_{j_t}$, where $\gamma_t:=\frac{2}{t+2}$. Let $\bx^*$ be a minimizer of $F$ over $\Delta(S)$. Then, for every integer $T\ge 0$:
    \begin{enumerate}[label=(\roman*)]
    	\item $\bx^{(T)}$ uses at most $T+1$ alternatives;
    	\item $F(\bx^{(T)}) - F(\bx^*) \le \frac{4L_p}{T+2}$;
        \item $V_p(\bx^{(T)}) \le \left( V_p(\bx^*)^p + \frac{4L_p}{T+2}\right)^{1/p}
    \le V_p(\bx^*) + \left(\frac{4L_p}{T+2}\right)^{1/p}$. If, moreover, $V_p(\bx^*)>0$, then $V_p(\bx^{(T)}) - V_p(\bx^*)
    \le \frac{4L_p}{p(T+2)V_p(\bx^*)^{p-1}}
    = \frac{4(p-1)2^{(p-2)/2}}{(T+2)V_p(\bx^*)^{p-1}}$.
    \end{enumerate}
\end{theorem}

\begin{proof}
    Extend $F$ to $\mathbb R^m$ by the same formula,
    \[
    F(\bx):=\frac1n\sum_{i\in N}(\max\{0,\hat\tau_i-\langle \hat u_i,\bx\rangle\})^p,
    \quad \bx\in\mathbb R^m.
    \]
    For $p>1$, the scalar function $z\mapsto (\max\{0,z\})^p$ is differentiable on $\mathbb R$, with derivative
    $p(\max\{0,z\})^{p-1}$. 
    Thus, this ambient extension is differentiable on $\mathbb R^m$, and for every $\bx\in\Delta(S)$,
    \[
    \nabla F(\bx)=-\frac{p}{n}\sum_{i\in N} v_i(\bx)^{p-1}\hat u_i.
    \]
    Therefore, each step of the Frank--Wolfe method chooses a pure lottery $\be_j$ minimizing $\langle \nabla F(\bx), \be_j\rangle$, or equivalently maximizing $\sum_{i\in N} v_i(\bx)^{p-1}\hat u_i(s_j)$.
    
    We next prove that $\nabla F$ is $L_p$-Lipschitz on $\Delta(S)$. Fix $\bx,\mathbf{y}\in\Delta(S)$ and $i\in N$. Since $v_i\le\sqrt{2}$ on $\Delta(S)$, the derivative of $z\mapsto z^{p-1}$ is at most $(p-1)(\sqrt{2})^{p-2}$ on $[0,\sqrt{2}]$, and $|v_i(\bx)-v_i(\mathbf{y})|\le|\langle \hat u_i,\bx-\mathbf{y}\rangle|$, we have
    \[
    |v_i(\bx)^{p-1} - v_i(\mathbf{y})^{p-1}|
    \le
    (p-1)(\sqrt{2})^{p-2} |\langle \hat u_i,\bx-\mathbf{y}\rangle|.
    \]
    Therefore,
    \[
    \|\nabla F(\bx)-\nabla F(\mathbf{y})\|
    =
    \left\|
    \frac{p}{n}\sum_{i\in N}(v_i(\bx)^{p-1}-v_i(\mathbf{y})^{p-1})\hat u_i
    \right\|
    \le
    \frac{p}{n}\sum_{i\in N}|v_i(\bx)^{p-1}-v_i(\mathbf{y})^{p-1}|\,\|\hat u_i\|
    \]
    \[
    \le
    p(p-1)(\sqrt{2})^{p-2}\frac1n \sum_{i\in N} |\langle \hat u_i,\bx-\mathbf{y}\rangle|
    \le
    p(p-1)(\sqrt{2})^{p-2}\|\bx-\mathbf{y}\|
    =
    L_p\|\bx-\mathbf{y}\|.
    \]
    
    Let $h_t := F(\bx^{(t)}) - F(\bx^*)$.
    Since $\nabla F$ is $L_p$-Lipschitz on the convex set $\Delta(S)$, the standard smoothness inequality gives
    \[
    F((1-\gamma)\bx+\gamma \mathbf{z})
    \le
    F(\bx) + \gamma \langle \nabla F(\bx), \mathbf{z} - \bx\rangle + \frac{L_p}{2}\gamma^2 \|\mathbf{z} - \bx\|^2
    \]
    for all $\bx, \mathbf{z}\in\Delta(S)$ and all $\gamma\in[0,1]$. 
    Applying this with $\bx=\bx^{(t)}$, $\mathbf{z}=\be_{j_t}$, and $\gamma=\gamma_t$, and using $\|\be_{j_t}-\bx^{(t)}\|\le \sqrt{2}$, gives
    \[
    F(\bx^{(t+1)})
    \le F(\bx^{(t)}) + \gamma_t \langle \nabla F(\bx^{(t)}), \be_{j_t} - \bx^{(t)}\rangle + L_p \gamma_t^2.
    \]
    By the choice of $\be_{j_t}$,
    \[
    \langle \nabla F(\bx^{(t)}), \be_{j_t} - \bx^{(t)}\rangle
    \le \langle \nabla F(\bx^{(t)}), \bx^* - \bx^{(t)}\rangle.
    \]
    Since $F$ is convex,
    \[
    \langle \nabla F(\bx^{(t)}), \bx^* - \bx^{(t)}\rangle
    \le
    F(\bx^*) - F(\bx^{(t)})
    = -h_t.
    \]
    Thus, $h_{t+1} \le (1-\gamma_t)h_t + L_p \gamma_t^2$.
    With $\gamma_t = 2/(t+2)$, we prove by induction that
    \[
    h_t \le \frac{4L_p}{t+2}
    \quad\text{for all } t\ge 0.
    \]
    For $t=0$, since every $v_i(\bx^{(0)}) \le \sqrt{2}$,
    \[
    h_0 \le F(\bx^{(0)}) \le (\sqrt{2})^p \le 2L_p = \frac{4L_p}{2},
    \]
    because $p(p-1)\ge 1$ for $p\ge 2$. Now assume $h_t \le 4L_p/(t+2)$. Then
    \[
    h_{t+1}
    \le
    \left(1-\frac{2}{t+2}\right)\frac{4L_p}{t+2} + L_p\frac{4}{(t+2)^2}
    =
    \frac{4L_p(t+1)}{(t+2)^2}
    \le
    \frac{4L_p}{t+3},
    \]
    which closes the induction. This proves (ii).
    
    Part~(i) holds because $\bx^{(0)}$ is a pure lottery and each update writes $\bx^{(t+1)}$ as a convex combination of $\bx^{(t)}$ and one pure lottery $\be_{j_t}$, so each step adds at most one alternative.
    
    For (iii), use $F(\bx)=V_p(\bx)^p$:
    \[
    V_p(\bx^{(T)})^p - V_p(\bx^*)^p
    =
    F(\bx^{(T)}) - F(\bx^*)
    \le
    \frac{4L_p}{T+2}.
    \]
    Consequently,
    \[
    V_p(\bx^{(T)})
    \le \left(
    V_p(\bx^*)^p + \frac{4L_p}{T+2}
    \right)^{1/p}.
    \]
    The displayed additive bound by $(4L_p/(T+2))^{1/p}$ follows from $(a+b)^{1/p} \le a^{1/p} + b^{1/p}$ for $a,b\ge 0$.
    If $V_p(\bx^*)>0$, then the function $r\mapsto r^{1/p}$ is concave on $\mathbb{R}_{\ge 0}$, so
    \[
    (a+b)^{1/p} - a^{1/p}
    \le
    \frac{1}{p}a^{1/p-1}b
    \quad \text{for all } a>0 \text{ and } b\ge 0.
    \]
    Applying this with $a=V_p(\bx^*)^p$ and $b=4L_p/(T+2)$ gives
    \[
    V_p(\bx^{(T)}) - V_p(\bx^*)
    \le
    \frac{4L_p}{p(T+2)V_p(\bx^*)^{p-1}}
    =
    \frac{4(p-1)2^{(p-2)/2}}{(T+2)V_p(\bx^*)^{p-1}},
    \]
    as claimed.
\end{proof}

\section{Omitted Proofs and Details from Section~\ref{sec:budget}}
  
\subsection{Proof of Theorem~\ref{thm:alg_error_bound}}
    By Step~1 of Algorithm~\ref{alg:budgeted-fixed-p}, $kT(\eta)\le Q$, and each of the $k$ invocations of the subroutine uses at most $T(\eta)$ membership queries. 
    Thus, Algorithm~\ref{alg:budgeted-fixed-p} uses at most $Q$ membership queries. Let $\hat \bx$ denote the lottery returned by Algorithm~\ref{alg:budgeted-fixed-p}, and let $\bx^*$ be a minimizer of $F$ over $\Delta(S)$. We compare $F$ with the objective of the sampled agents and with the objective that the algorithm minimizes, namely,
    \[
    F_k(\bx):=\frac{1}{k}\sum_{t=1}^k v_{i_t}(\bx)^p
    \quad\text{and}\quad
    \widetilde F_k(\bx):=\frac{1}{k}\sum_{t=1}^k \tilde v_t(\bx)^p,
    \]
    where $\tilde v_t$ is defined in Algorithm~\ref{alg:budgeted-fixed-p}. The difference between $\widetilde F_k$ and $F_k$ comes from boundary learning, and the difference between $F_k$ and $F$ comes from sampling.
    By the guarantee of $\LearnHalfspace$, for each $t\in[k]$,
    \[
    \|\tilde u_t-\hat u_{i_t}\|\le \eta,
    \quad
    |\tilde\tau_t-\hat\tau_{i_t}|\le \eta.
    \]
    Since $z\mapsto \max\{0,z\}$ is $1$-Lipschitz and $\|\bx\|\le 1$ for all $\bx\in\Delta(S)$,
    for every $\bx\in\Delta(S)$,
    \begin{equation*}
        |\tilde v_t(\bx)-v_{i_t}(\bx)|
        \le
        |(\tilde\tau_t-\langle \tilde u_t,\bx\rangle)
        -(\hat\tau_{i_t}-\langle \hat u_{i_t},\bx\rangle)|
        \le
        |\tilde\tau_t-\hat\tau_{i_t}|
        +\|\tilde u_t-\hat u_{i_t}\|\,\|\bx\|
        \le 2\eta .
    \end{equation*}
    Since $v_i\le\sqrt{2}$ on $\Delta(S)$ for every agent $i$ (Section~\ref{sec:preliminaries}), it follows that $\widetilde v_t(\bx)\le v_{i_t}(\bx)+2\eta\le \sqrt{2}+2\eta$ for every $t\in[k]$ and $\bx\in\Delta(S)$.
    
    Since $\tilde v_t(\bx), v_{i_t}(\bx) \in [0,\sqrt{2}+2\eta]$ with $p\ge 1$, the mean value theorem gives, uniformly over $\bx\in\Delta(S)$,
    \begin{align*}
        |\tilde v_t(\bx)^p-v_{i_t}(\bx)^p|
        \le p(\sqrt{2}+2\eta)^{p-1}\,|\tilde v_t(\bx)-v_{i_t}(\bx)|
        \le 2p\eta(\sqrt{2}+2\eta)^{p-1}.
    \end{align*}
    Averaging over $t$ gives
    \begin{equation}\label{eq:plugin-gap}
        \sup_{\bx\in\Delta(S)}|\widetilde F_k(\bx)-F_k(\bx)|
        \le 2p\eta (\sqrt2+2\eta)^{p-1}.
    \end{equation}
    
    Then, 
    \begin{align*}
        F(\hat \bx)-F(\bx^*)
        &=
        (F(\hat \bx)-F_k(\hat \bx))
        +(F_k(\hat \bx)-\widetilde F_k(\hat \bx))
        + (\widetilde F_k(\hat \bx)-\widetilde F_k(\bx^*)) \\
        &\,\quad
        +(\widetilde F_k(\bx^*)-F_k(\bx^*)) +(F_k(\bx^*)-F(\bx^*)) \\
        &\le |F(\hat \bx)-F_k(\hat \bx)|
        +|F_k(\hat \bx)-\widetilde F_k(\hat \bx)|
        +|\widetilde F_k(\bx^*)-F_k(\bx^*)|
        +|F_k(\bx^*)-F(\bx^*)| \\
        & \le
        2\sup_{\bx\in\Delta(S)}|F(\bx)-F_k(\bx)|
        +2\sup_{\bx\in\Delta(S)}|\widetilde F_k(\bx)-F_k(\bx)|.
    \end{align*}
    Combining this with~\eqref{eq:plugin-gap}, we obtain
    \begin{equation}\label{eq:main-decomp}
        F(\hat \bx)\le F(\bx^*)
        +2\sup_{\bx\in\Delta(S)}|F(\bx)-F_k(\bx)|
        +4p\eta(\sqrt2+2\eta)^{p-1}.
    \end{equation}
    
    Let $\sigma_1,\dots,\sigma_k$ be i.i.d.\ Rademacher signs (uniform on $\{\pm1\}$),
    independent of $i_1,\dots,i_k$.
    Standard symmetrization gives
    \begin{equation}\label{eq:sym}
        \mathbb{E}\left[\sup_{\bx\in\Delta(S)}\left(F(\bx)-F_k(\bx)\right)\right]
        \le
        \frac{2}{k}\,
        \mathbb{E}\left[\sup_{\bx\in\Delta(S)}\sum_{t=1}^k \sigma_t  v_{i_t}(\bx)^p\right].
    \end{equation}
    Since $v_{i_t}(\bx)=\max \{0, \hat\tau_{i_t}-\langle \hat u_{i_t},\bx\rangle \} \le \sqrt{2}$ for all $\bx\in\Delta(S)$,
    the function $z \mapsto \max \{0, z\}^p$ is $p(\sqrt{2})^{p-1}$-Lipschitz on the set $\{\hat\tau_{i_t}-\langle \hat u_{i_t},\bx\rangle : \bx\in\Delta(S)\}\subseteq(-\infty,\sqrt{2}]$.
    By the contraction inequality for Rademacher averages,
    \begin{align*}
        \mathbb{E}\left[\sup_{\bx\in\Delta(S)}\sum_{t=1}^k \sigma_t\, v_{i_t}(\bx)^p\right]
        &\le
        p(\sqrt2)^{p-1}\,
        \mathbb{E}\left[\sup_{\bx\in\Delta(S)}\sum_{t=1}^k \sigma_t\,(\hat\tau_{i_t}-\langle \hat u_{i_t},\bx\rangle)\right] \\
        &=
        p(\sqrt2)^{p-1}\,
        \mathbb{E}\left[\sup_{\bx\in\Delta(S)}\left\langle -\sum_{t=1}^k \sigma_t \hat u_{i_t}, \bx\right\rangle\right],
    \end{align*}
    where the equality holds because $\sum_{t=1}^k\sigma_t\hat\tau_{i_t}$ does not depend on $\bx$ and has zero expectation over $(\sigma_t)$.
    By the Cauchy--Schwarz inequality and since $\|\bx\|\le 1$ for all $\bx\in\Delta(S)$,
    \begin{equation*}
        \sup_{\bx\in\Delta(S)}\left\langle -\sum_{t=1}^k \sigma_t \hat u_{i_t}, \bx\right\rangle
        \le
        \left\|\sum_{t=1}^k \sigma_t \hat u_{i_t}\right\|.
    \end{equation*}
    Finally, since $\|\hat u_i\|=1$ for all $i$,
    \begin{equation*}
        \mathbb{E}\left[\left\|\sum_{t=1}^k \sigma_t \hat u_{i_t}\right\|\right]
        \le
        \sqrt{\mathbb{E}\left[\left\|\sum_{t=1}^k \sigma_t \hat u_{i_t}\right\|^2\right]}
        =
        \sqrt{\sum_{t=1}^k \mathbb{E}[\|\hat u_{i_t}\|^2]}
        =
        \sqrt{k}.
    \end{equation*}
    Combining this with~\eqref{eq:sym} gives
    \begin{equation}\label{eq:exp-one-tail}
        \mathbb{E}\left[\sup_{\bx\in\Delta(S)}\left(F(\bx)-F_k(\bx)\right)\right]
        \le \frac{2p(\sqrt2)^{p-1}}{\sqrt{k}}.
    \end{equation}
    By the same argument, the same bound holds for $\mathbb{E}[\sup_{\bx \in \Delta(S)}(F_k(\bx)-F(\bx))]$.
    
    Next, we apply McDiarmid's inequality. 
    For any fixed $\bx\in\Delta(S)$, $F_k(\bx)$ is an average
    of $k$ numbers in $[0,(\sqrt2)^p]$, so changing one sampled index $i_t$ can change $F_k(\bx)$ by at most $(\sqrt2)^p/k$. Thus, changing one $i_t$ can change $\sup_{\bx\in\Delta(S)}(F(\bx)-F_k(\bx))$ by at most $(\sqrt2)^p/k$ as well.
    Therefore, with probability at least $1-\delta/2$,
    \begin{equation*}
        \sup_{\bx\in\Delta(S)}(F(\bx)-F_k(\bx))
        \le
        \mathbb{E}\left[\sup_{\bx\in\Delta(S)}(F(\bx)-F_k(\bx))\right]
        +(\sqrt2)^p\sqrt{\frac{\log(2/\delta)}{2k}}.
    \end{equation*}
    Applying the same argument to $\sup_{\bx \in \Delta(S)}(F_k(\bx)-F(\bx))$ and taking a union bound over the two events, with probability at least $1-\delta$ we have
    \begin{equation}\label{eq:uniform-gap}
        \sup_{\bx\in\Delta(S)}|F(\bx) - F_k(\bx)|
        \le \frac{2p(\sqrt2)^{p-1}}{\sqrt{k}}
        +(\sqrt2)^p\sqrt{\frac{\log(2/\delta)}{2k}}.
    \end{equation}
    
    By the definition of $\rho_k(\delta)$, inequality~(\ref{eq:uniform-gap}) states that $\sup_{\bx\in\Delta(S)} |F(\bx)-F_k(\bx)| \le \rho_k(\delta)$ with probability at least $1-\delta$. Substituting this into (\ref{eq:main-decomp}), we obtain
    \begin{equation*}
        F(\hat{\bx}) \le F(\bx^*) + 4p\eta(\sqrt{2}+2\eta)^{p-1} +
        2\rho_k(\delta).
    \end{equation*}
    Since $V_p(\bx)=F(\bx)^{1/p}$, we obtain
    \[
    V_p(\hat \bx)
    = F(\hat \bx)^{1/p}
    \le (F(\bx^*)+4p\eta(\sqrt{2}+2\eta)^{p-1}+2\rho_k(\delta))^{1/p}.
    \]
    Therefore, using $(a+b)^{1/p}\le a^{1/p}+b^{1/p}$ for $a,b\ge 0$,
    \begin{align*}
    V_p(\hat \bx)
    & \le V_p(\bx^*)+(4p\eta(\sqrt{2}+2\eta)^{p-1}+2\rho_k(\delta))^{1/p} \\
    & = \min_{\bx\in\Delta(S)} V_p(\bx)+(4p\eta(\sqrt{2}+2\eta)^{p-1}+2\rho_k(\delta))^{1/p}.
    \end{align*}
    This proves the first expression in the $\min$ term.
    
    For the second expression in the $\min$ term, define
    \[
    G_k(\bx) := F_k(\bx)^{1/p}
    = \left(\frac{1}{k}\sum_{t=1}^k v_{i_t}(\bx)^p\right)^{1/p},
    \quad
    \widetilde G_k(\bx) := \widetilde F_k(\bx)^{1/p}
    = \left(\frac{1}{k}\sum_{t=1}^k \widetilde v_t(\bx)^p\right)^{1/p}.
    \]
    Since $r\mapsto r^{1/p}$ is strictly increasing on $\mathbb{R}_{\ge 0}$ and $\hat \bx\in\arg\min_{\bx\in\Delta(S)} \widetilde F_k(\bx)$, we also have $\hat \bx\in\arg\min_{\bx\in\Delta(S)} \widetilde G_k(\bx)$.
    Since $|\widetilde v_t(\bx)-v_{i_t}(\bx)|\le 2\eta$ for every $t\in[k]$ and every $\bx\in\Delta(S)$, the reverse triangle inequality for the $\ell_p$-norm gives, uniformly over $\bx\in\Delta(S)$,
    \begin{equation*}
        |\widetilde{G}_k(\bx)-G_k(\bx)|
        \le \left( \frac{1}{k}\sum_{t=1}^k |\widetilde{v}_{t}(\bx)-v_{i_t}(\bx)|^p \right)^{1/p}
        \le 2\eta.
    \end{equation*}
    Moreover, on the same event as above, for every $\bx\in\Delta(S)$,
    \begin{equation*}
        |V_p(\bx)-G_k(\bx)|
        = |F(\bx)^{1/p} - F_k(\bx)^{1/p}|
        \le |F(\bx)-F_k(\bx)|^{1/p}
        \le \rho_k(\delta)^{1/p},
    \end{equation*}
    where we used that $z\mapsto z^{1/p}$ is $1/p$-H\"older on $\mathbb{R}_{\ge 0}$.
    
    Therefore, on this event,
    \begin{align*}
        V_p(\hat{\bx}) \le G_k(\hat{\bx}) + \rho_k(\delta)^{1/p} 
        &\le \widetilde{G}_k(\hat{\bx}) + 2\eta + \rho_k(\delta)^{1/p} \\
        &\le \widetilde{G}_k(\bx^*) + 2\eta + \rho_k(\delta)^{1/p} \\
        &\le G_k(\bx^*) + 4\eta + \rho_k(\delta)^{1/p} \\
        &\le V_p(\bx^*) + 4\eta + 2\rho_k(\delta)^{1/p} = \min_{\bx\in\Delta(S)} V_p(\bx) + 4\eta + 2\rho_k(\delta)^{1/p}.
    \end{align*}
    This proves the second expression. Taking the minimum of the expressions completes the proof.    

\subsection{Proof of Corollary~\ref{cor:main_guarantee}}
    If $\alpha \ge \sqrt{2}$, output any lottery $\bx$ without making queries. Since $v_i\le\sqrt{2}$ on $\Delta(S)$ for every agent $i$, we have $V_p(\bx) \le \sqrt{2} \le \min_{\mathbf{y}\in\Delta(S)} V_p(\mathbf{y}) + \alpha$.

    Now assume that $\alpha < \sqrt{2}$, and run Algorithm~\ref{alg:budgeted-fixed-p} with $\eta=0$, using the routine of Lemma~\ref{lem:learn-halfspace}, which makes $\mathcal{O}(m\log(1/\varepsilon))$ queries per agent. By Theorem~\ref{thm:alg_error_bound}, with probability at least $1-\delta$,
    \[
    V_p(\hat\bx)\le \min_{\mathbf{y}\in\Delta(S)} V_p(\mathbf{y})+(2\rho_k(\delta))^{1/p}.
    \]
    Since $1+\sqrt{z}\le\sqrt{2(1+z)}$ for $z\ge 0$, there is a constant $c_p>0$, depending only on $p$, such that $\rho_k(\delta)\le c_p\sqrt{(1+\log(2/\delta))/k}$. Hence choosing $k=\lceil 4c_p^2(1+\log(2/\delta))\alpha^{-2p}\rceil$ ensures that $(2\rho_k(\delta))^{1/p}\le\alpha$. Since $\alpha<\sqrt{2}$ and $\delta<1$, the quantity $(1+\log(2/\delta))\alpha^{-2p}$ is bounded below by a positive constant depending only on $p$, so the rounding up only affects the constant. Therefore the total number of queries is $k\cdot\mathcal{O}(m\log(1/\varepsilon))\le C_p\, m \log(1/\varepsilon)\, (1+\log(2/\delta))\, \alpha^{-2p}$ for a constant $C_p > 0$ depending only on $p$.

\subsection{Proof of Theorem~\ref{thm:unqueried-agents}}\label{app:proof-unqueried}
    Let $m=3$, and first consider the profile in which every agent has $u_i=(1,1,0)$ and $\tau_i = 1$. For every $\bx \in \Delta(S)$, we have $U_i(\bx) = x_1+x_2 = 1-x_3$, so agent $i$ accepts $\bx$ if and only if $x_3=0$. Let $N'$ be the set of agents that the algorithm queries on this profile, so that $|N'|\le K$, and let $\hat\bx$ be its output. Choose $r \in \argmin_{j\in\{1,2\}} \hat{x}_j$. Since $\hat{x}_1+\hat{x}_2 \le 1$, we have $\hat{x}_r \le \tfrac12$.

    Now change only the agents who are not queried: each agent $i\in N\setminus N'$ now has $u_i=\be_r$ and $\tau_i=1$, while each agent in $N'$ keeps $u_i=(1,1,0)$ and $\tau_i=1$. In the new profile, every agent accepts some lottery and rejects another: agents in $N'$ accept $\be_1$ and $\be_2$ and reject $\be_3$, and agents in $N\setminus N'$ accept $\be_r$ and reject $\be_3$. Every query that the algorithm makes on the original profile is to an agent in $N'$, whose answers do not change. Since the algorithm is deterministic, an induction on the number of queries shows that it makes the same queries, receives the same answers, and returns the same lottery $\hat\bx$ on the new profile.

    In the new profile, every agent accepts $\be_r$: the agents in $N'$ because its third coordinate is $0$, and the agents in $N\setminus N'$ because $U_i(\be_r)=1=\tau_i$. Hence $\min_{\bx \in \Delta(S)} V_p(\bx)=0$. On the other hand, fix any $i \in N \setminus N'$. Then $\bar u_i=\frac13$, and $w_i$ has entry $\frac23$ at coordinate $r$ and $-\frac13$ at the other two coordinates, so $\|w_i\| = \sqrt{6}/3$. Since $U_i(\hat\bx)=\hat x_r\le\frac12$, we have
    \[
    v_i(\hat{\bx}) = \frac{1-\hat{x}_r}{\sqrt{6}/3} \ge \frac{1/2}{\sqrt{6}/3} = \sqrt{\frac38}.
    \]
    Consequently,
    \[
    V_p(\hat{\bx})^p \ge \frac{|N \setminus N'|}{n} \left(\frac38\right)^{p/2} \ge \frac{n-K}{n} \left(\frac38\right)^{p/2},
    \]
    and taking $p$-th roots gives $V_p(\hat\bx)\ge\sqrt{3/8}\,((n-K)/n)^{1/p}$, which proves the bound. In particular, a deterministic algorithm that always outputs a minimizer of $V_p$ cannot query at most $K\le n-1$ agents on every profile.

    For the second claim, repeat the argument with threshold $\frac12$ instead of $1$ for all agents in the initial profile, so that each of them accepts $\bx$ if and only if $x_3\le\frac12$, and with threshold $\frac34$ instead of $1$ for the changed agents. Then every agent strictly accepts some pure alternative and rejects $\be_3$, every agent still accepts $\be_r$, and each changed agent has violation $(\frac34-\hat x_r)/(\sqrt{6}/3)\ge\frac{3}{4\sqrt{6}}=\frac{\sqrt{6}}{8}$ at $\hat\bx$. The rest of the argument is unchanged.

\subsection{Randomized Algorithms that Always Output a Minimizer}\label{app:randomized-exact}

\begin{proposition}\label{prop:randomized-exact}
    Let $m\ge 3$ and $p\in[1,\infty)$, and consider a randomized, possibly adaptive, algorithm that outputs a minimizer of $V_p$ with probability one on every profile. Then, for every agent $i$, there is a profile in which every agent strictly accepts some pure alternative and on which the algorithm queries agent $i$ with probability one.
\end{proposition}

\begin{proof}
    Fix $i\in N$. Consider two profiles in which every agent other than $i$ has utility vector $\be_1+\be_2$ and threshold $\frac12$, and therefore accepts exactly the lotteries with $x_1+x_2\ge\frac12$. In the first profile, agent $i$ has $u_i=\be_1$ and $\tau_i=\frac34$; in the second profile, she has $u_i=\be_2$ and $\tau_i=\frac34$. In both profiles, every agent strictly accepts some pure alternative and rejects $\be_3$. In the first profile, the lotteries accepted by all agents are those with $x_1\ge\frac34$; these lotteries have $V_p=0$, so they are exactly the minimizers of $V_p$. Similarly, in the second profile, the minimizers of $V_p$ are the lotteries with $x_2\ge\frac34$. No lottery belongs to both sets.

    Fix a realization of the internal randomness of the algorithm for which the algorithm does not query agent $i$ on the first profile. Then all of its queries on the first profile are to agents other than $i$, whose answers are the same in both profiles, so the algorithm makes the same queries and returns the same lottery on both profiles. Since no lottery minimizes $V_p$ in both profiles, this lottery is not a minimizer on one of the two profiles. As the algorithm outputs a minimizer with probability one on each profile, such realizations have probability zero. Hence the algorithm queries agent $i$ with probability one on the first profile.
\end{proof}

\subsection{Upper Bound for Unqueried Agents}\label{app:coverage-only-upper}

This subsection shows that the dependence on $(n-K)/n$ in Theorem~\ref{thm:unqueried-agents} cannot be improved. Fix a set $N'\subseteq N$ of $K$ agents whose normalized boundaries are known, and minimize their contribution $\frac1n\sum_{i\in N'}v_i(\bx)^p$ to $F$. This removes boundary-learning error and sampling error, so the only remaining loss comes from the agents in $N\setminus N'$.

\begin{proposition}\label{prop:coverage_upper}
    Fix $p \in [1,\infty)$, let $N' \subseteq N$ be a set of $K$ agents whose normalized boundaries are known, and let $\hat \bx$ minimize $\frac{1}{n}\sum_{i\in N'} v_i(\bx)^p$ over $\Delta(S)$. Then
    \[
    V_p(\hat \bx) \le \min_{\bx\in\Delta(S)} V_p(\bx) + \sqrt{2}\left(\frac{n-K}{n}\right)^{1/p}.
    \]
    In particular, if every agent in $N'$ strictly accepts some pure alternative, there is a deterministic algorithm that only queries the agents in $N'$, makes $\mathcal{O}(Km\log(1/\varepsilon))$ queries on every $\varepsilon$-quantized instance, and satisfies the same guarantee.
\end{proposition}

\begin{proof}
    A minimizer $\hat\bx$ exists by continuity and compactness; we keep the factor $1/n$ so that the restricted objective is on the same scale as $F$. Let $\bx^*$ minimize $V_p$. By the choice of $\hat\bx$,
    \[
    V_p(\hat \bx)^p = \frac{1}{n}\sum_{i\in N'} v_i(\hat \bx)^p + \frac{1}{n}\sum_{i\in N\setminus N'} v_i(\hat \bx)^p \le V_p(\bx^*)^p + \frac{1}{n}\sum_{i\in N\setminus N'} v_i(\hat \bx)^p.
    \]
    Since $v_i\le\sqrt{2}$ on $\Delta(S)$ for every agent $i$, the last sum is at most $\frac{n-K}{n}(\sqrt{2})^p$. Taking $p$-th roots and using $(a+b)^{1/p}\le a^{1/p}+b^{1/p}$ for $a,b\ge 0$ gives the bound. For the last claim, Lemma~\ref{lem:learn-halfspace} finds the normalized boundary of each agent in $N'$ with $\mathcal{O}(m\log(1/\varepsilon))$ queries.
\end{proof}

\subsection{Proof of Proposition~\ref{prop:noise}}\label{app:proof-noise}

Since the modified algorithm stops after $L$ queries and asks each of them $r$ times, it makes at most $rL$ queries. Fix the internal randomness of the original algorithm, if any. As long as all majority answers are correct, the modified algorithm behaves exactly as the original algorithm with exact answers, and in particular, it finishes within $L$ queries. For $t\in[L]$, let $E_t$ be the event that the modified algorithm asks a $t$-th query, the majority answers to its first $t-1$ queries are correct, and the majority answer to its $t$-th query is wrong. If none of these events occurs, then all majority answers are correct.

Fix $t\in[L]$. Conditioned on everything that happens before the $t$-th query, the $r$ answers to this query are flipped independently, each with probability at most $\nu$. Let $X$ be the number of flipped answers, so that $\mathbb{E}[X]\le\nu r$. Since $r$ is odd, the majority answer is wrong only if $X\ge r/2$, and Hoeffding's inequality gives
\begin{align*}
\Pr\left[X\ge \frac r2\right]
&\le\Pr\left[X-\mathbb{E}[X]\ge\left(\frac12-\nu\right)r\right]\\
&\le\exp\left(-2r\left(\frac12-\nu\right)^2\right)=\exp\left(-\frac{r(1-2\nu)^2}{2}\right)\le\frac{\delta}{L},
\end{align*}
where the last inequality follows from the choice of $r$. Hence $\Pr[E_t]\le\delta/L$, and by the union bound, the probability that some majority answer is wrong is at most $\sum_{t=1}^L\Pr[E_t]\le\delta$.

\subsection{Weighted Agents}\label{app:weights}

Let $a_i\ge 0$ for each $i\in N$ with $\sum_{i\in N}a_i=1$, fix $p\in[1,\infty)$, and consider the weighted objective $V_{p,a}(\bx):=\big(\sum_{i\in N}a_iv_i(\bx)^p\big)^{1/p}$, which equals $V_p$ when $a_i=1/n$ for all $i$. We explain how our proofs change.

\textbf{Theorem~\ref{thm:worst-case-objective}.} Replace $\bar F$ by $\bx\mapsto\sum_{i\notin B}a_iv_i(\bx)^p+\sum_{i\in B}a_i\,\mathrm{dist}(\bx,A_i)^p$. The proof goes through with $\frac1n$ replaced by $a_i$, and gives $V_{p,a}(\bar\bx)^p\le V_{p,a}(\bx^*)^p+(\sqrt{2})^p\sum_{i\in B}a_i$, where $\bx^*$ minimizes $V_{p,a}$. Hence $V_{p,a}(\bar\bx)\le V_{p,a}(\bx^*)+\sqrt{2}\,(\sum_{i\in B}a_i)^{1/p}$.

\textbf{Theorem~\ref{thm:dual-finite-p}.} For $1<p<\infty$, the function $\bx\mapsto\sum_{i\in N}a_iv_i(\bx)^p$ is convex and differentiable (see the proof of Theorem~\ref{thm:fw-sparse}), with gradient $-p\sum_{i\in N}a_iv_i(\bx)^{p-1}\hat u_i$. A lottery minimizes a convex differentiable function over $\Delta(S)$ if and only if every alternative that it uses minimizes the corresponding coordinate of the gradient. Hence $\bx^*$ minimizes $V_{p,a}$ if and only if every alternative that $\bx^*$ uses maximizes $\sum_{i\in N}a_iv_i(\bx^*)^{p-1}\hat u_i(s_j)$.

\textbf{Theorem~\ref{thm:alg_error_bound}.} Suppose that Algorithm~\ref{alg:budgeted-fixed-p} samples $i_1,\dots,i_k$ independently from the distribution $(a_i)_{i\in N}$. Then, for every $\bx$, the expectation of $F_k(\bx)=\frac1k\sum_{t=1}^k v_{i_t}(\bx)^p$ is $\sum_{i\in N}a_iv_i(\bx)^p$. The proof of Theorem~\ref{thm:alg_error_bound} only uses that the sampled agents are independent and identically distributed and that the expectation of $F_k$ equals the objective, so the theorem holds unchanged with $V_p$ replaced by $V_{p,a}$.

\section{Experimental Details}
\label{app:experiments}

This appendix describes the populations, the learner, and the solver used in Section~\ref{sec:experiments}, together with additional experiments. Our code, together with the saved results and instructions for reproducing the experiments with a single command, is available at \codeurl. Throughout, the algorithm only observes membership query answers of the sampled agents; the true boundaries of all agents are used only afterwards, to evaluate the computed lotteries.

\paragraph{Populations.}
Every agent is generated in the same way. We draw a preference direction by adding independent Gaussian noise to each coordinate of the center of the agent's cluster, and rescale it affinely to $[0,1]$, so that the agent's best alternative has utility $1$ and her worst alternative has utility~$0$. Her threshold is $\tau_i=\alpha_i$, where $\alpha_i$ is drawn from a Gaussian distribution whose mean is the threshold offset of her cluster and whose standard deviation is $0.035$, and then clipped to $[0.2,0.9]$. Thus the threshold lies a fraction $\alpha_i$ of the way from the lowest to the highest utility of the agent, and larger offsets mean stricter agents. To make the instance $\varepsilon$-quantized with $\varepsilon=1/32$, we round each utility to the nearest multiple of $1/32$, keeping the best and worst utilities at $1$ and $0$, and move the threshold to the nearest multiple of $1/32$ strictly between $0$ and $1$ that is different from all utilities of the agent. Hence every agent strictly accepts her best pure alternative and rejects her worst one, so no agent is distance-underdetermined. Table~\ref{tab:finite-query-setup} lists the parameters of the two populations.

\begin{table*}[h]
\centering
\scriptsize
\setlength{\tabcolsep}{4pt}
\begin{tabular}{>{\raggedright\arraybackslash}p{2.2cm}>{\raggedright\arraybackslash}p{5.4cm}>{\raggedright\arraybackslash}p{5.4cm}}
\toprule
& Shared-block population & Conflicting-blocks population \\
\midrule
$(n,m)$ & $(1000,10)$ & $(1000,10)$ \\
Clusters & 3 clusters with probabilities $0.36/0.34/0.30$. & 5 clusters with probabilities $0.30/0.24/0.21/0.17/0.08$. \\
Cluster centers & The common pattern $(1.2,1.0,0.8,0,\dots,0)$ plus independent Gaussian noise with standard deviation $0.22$ in each coordinate, drawn once for each cluster of each population. & Cluster $b\in\{1,\dots,5\}$ has value $1.25$ on alternatives $2b-1$ and $2b$, and value $-0.42$ on the other alternatives. \\
Agent noise & Standard deviation $0.28$. & Standard deviation $0.18$. \\
Threshold offsets & $0.70/0.75/0.72$. & $0.32/0.38/0.35/0.42/0.70$; the smallest cluster is the strictest. \\
Query budget & $Q=3200$; $k\in\{5,10,20,40,80,160,320\}$; $\lfloor Q/k\rfloor$ queries per sampled agent. & Same. \\
Populations & 100 populations, each with its own samples of agents. & Same. \\
\bottomrule
\end{tabular}
\caption{Setup of the finite-query experiments in Figures~\ref{fig:query-budget} and~\ref{fig:appendix-query-diagnostics}.}
\label{tab:finite-query-setup}
\end{table*}

\paragraph{Learner.}
The learner follows case~1 in the proof of Theorem~\ref{thm:identification} (Appendix~\ref{app:proof-identification}), which is also the procedure of Lemma~\ref{lem:learn-halfspace}. It queries the $m$ pure lotteries and fixes the rejected alternative $b$ with the smallest index. It then searches for the boundary point on the edge from $\be_b$ to $\be_j$ for each accepted alternative $j$, lets $a$ be an accepted alternative whose boundary point is closest to $\be_b$, and searches for the boundary point on the edge from $\be_j$ to $\be_a$ for each rejected alternative $j\neq b$. Finally, it computes $(\hat u_i,\hat\tau_i)$ from these $m-1$ boundary points as at the end of the proof of Theorem~\ref{thm:identification}. Each search is a binary search for the parameter $\lambda^*$ of the boundary point. Since $\lambda^*$ is a fraction with denominator at most $32$, and any two such fractions differ by at least $1/32^2=2^{-10}$, after $11$ steps the search interval contains exactly one such fraction, which is $\lambda^*$. Hence $10+9\cdot 11=109$ queries recover the normalized boundary exactly. If an agent receives $d<109$ queries, the learner splits the $d-10$ queries left after the pure lotteries as evenly as possible among the $9$ searches, and each search with fewer than $11$ steps uses the midpoint of its final interval in place of $\lambda^*$. In particular, with $d=10$, the learner places every boundary point at the middle of its edge, and with $d=20$, one search takes two steps and the others take one step. If an agent receives more than $109$ queries, the learner only uses $109$ of them. Every query is answered by comparing the agent's utility for the queried lottery with her threshold in exact rational arithmetic, and is counted; the code checks that every run respects its budget, and that exact recovery returns the true normalized boundary up to floating-point error.

\paragraph{Objective and decomposition.}
Algorithm~\ref{alg:budgeted-fixed-p} samples $k$ agents uniformly at random with replacement. We compare three lotteries: the lottery $\bx_{\mathrm{full}}$ that minimizes $V_2$ for the full population, the lottery $\bx_{\mathrm{sample}}$ that minimizes the objective of the sampled agents with their true boundaries, and the output $\hat\bx$ of the algorithm, which minimizes the objective of the sampled agents with their learned boundaries. All three are evaluated with $V_2$ on the full population, which gives $G_{\mathrm{tot}}=V_2(\hat\bx)-V_2(\bx_{\mathrm{full}})$, $G_{\mathrm{cov}}=V_2(\bx_{\mathrm{sample}})-V_2(\bx_{\mathrm{full}})$, and $G_{\mathrm{bdry}}=V_2(\hat\bx)-V_2(\bx_{\mathrm{sample}})$, so that $G_{\mathrm{tot}}=G_{\mathrm{cov}}+G_{\mathrm{bdry}}$. The coverage gap is nonnegative, but the boundary-learning gap can be negative: a lottery computed from learned boundaries can happen to be better for the full population than one computed from the true boundaries of the same sampled agents. We minimize $F_2=V_2^2$, which has the same minimizers as $V_2$, and report all quantities in the scale of $V_2$. Algorithm~\ref{alg:budgeted-fixed-p} may return any minimizer, and the minimizer is not always unique; for example, when some lottery is acceptable to all sampled agents, every such lottery is a minimizer, which often happens for small $k$. Whenever we minimize $V_p$, both here and in the $p$-sweep below, we return the minimizer closest to the uniform lottery, that is, the minimizer with the smallest Euclidean norm. This choice affects the results for small $k$: another minimizer can give a smaller or a much larger coverage gap.

\paragraph{Solver.}
We minimize $V_p$ over the simplex with CVXPY~\citep{diamond2016cvxpy} and the interior-point solver Clarabel~\citep{goulart2024clarabel}: this is a quadratic program for $p=2$, a linear program for $p\in\{1,\infty\}$, and a conic program for other values of~$p$. To find the minimizer with the smallest norm, we first compute some minimizer $\bx'$, and then minimize $\|\bx\|^2$ over the lotteries $\bx$ with $v_i(\bx)\le v_i(\bx')$ for all agents $i$. For $1<p<\infty$, these are exactly the minimizers, since every minimizer gives each agent the same violation by the strict convexity of $t\mapsto t^p$ on $t>0$; for $p=1$ and $p=\infty$, we instead use the constraints $\sum_i v_i(\bx)\le \sum_i v_i(\bx')$ and $\max_i v_i(\bx)\le\max_i v_i(\bx')$, respectively. We set the duality gap and feasibility tolerances of Clarabel to $10^{-12}$, with at most $1000$ iterations. For some conic programs, Clarabel cannot reach this accuracy; we then retry with tolerances $10^{-10}$, $10^{-9}$, and $10^{-8}$, and only accept solutions that Clarabel reports as optimal.

\paragraph{Results.}
Table~\ref{tab:fixed-budget} lists the means shown in Figure~\ref{fig:query-budget}. For $k\le 20$, every sampled agent is learned exactly, and the boundary-learning gap is below $10^{-12}$ in every run. For $k\ge 40$, the budget of $Q=3200$ queries is used in full. To compare values of $k$, we pair the runs by population. For the conflicting-blocks population, the total excess at $k=80$ is smaller than at every other $k$, with 95\% bootstrap confidence intervals for the mean difference that exclude zero. For the shared-block population, the total excess at $k=80$ is significantly smaller than for $k\le 40$, but the confidence intervals for the differences from $k=160$ and $k=320$ contain zero.

\begin{table*}[h]
\centering
\small
\begin{tabular}{rrrrrrrr}
\toprule
& & \multicolumn{2}{c}{Shared-block} & \multicolumn{2}{c}{Conflicting-blocks} & \multicolumn{2}{c}{Middle thresholds} \\
\cmidrule(lr){3-4}\cmidrule(lr){5-6}\cmidrule(lr){7-8}
$k$ & $\lfloor Q/k\rfloor$ & $G_{\mathrm{tot}}$ & $G_{\mathrm{bdry}}$ & $G_{\mathrm{tot}}$ & $G_{\mathrm{bdry}}$ & $G_{\mathrm{tot}}$ & $G_{\mathrm{bdry}}$ \\
\midrule
$5$ & $640$ & $35.06$ & $0.00$ & $37.96$ & $0.00$ & $35.42$ & $0.00$ \\
$10$ & $320$ & $20.24$ & $0.00$ & $32.58$ & $0.00$ & $24.01$ & $0.00$ \\
$20$ & $160$ & $12.92$ & $0.00$ & $17.98$ & $0.00$ & $17.49$ & $0.00$ \\
$40$ & $80$ & $6.82$ & $0.01$ & $12.25$ & $-0.05$ & $11.78$ & $0.02$ \\
$80$ & $40$ & $3.30$ & $0.17$ & $6.02$ & $-0.06$ & $6.85$ & $-0.54$ \\
$160$ & $20$ & $3.90$ & $2.13$ & $10.28$ & $7.05$ & $12.00$ & $7.98$ \\
$320$ & $10$ & $4.24$ & $3.25$ & $37.29$ & $35.68$ & $6.31$ & $4.05$ \\
\bottomrule
\end{tabular}
\caption{Mean total excess and boundary-learning gap over 100 populations under a fixed budget of $Q=3200$ queries, in units of $10^{-3}$. The last two columns are for the conflicting-blocks population with thresholds closer to the middle of each agent's utility range.}
\label{tab:fixed-budget}
\end{table*}

\paragraph{Thresholds closer to the middle.}
The shape of the curves in Figure~\ref{fig:query-budget} depends on the population and on the learner. To illustrate this, we repeated the fixed-budget experiment for the conflicting-blocks population with threshold offsets $0.47/0.53/0.50/0.57/0.78$, so that the thresholds lie closer to the middle of each agent's utility range (last two columns of Table~\ref{tab:fixed-budget}). The total excess at $k=320$ is then $6.31\times 10^{-3}$, comparable to $6.85\times 10^{-3}$ at $k=80$ (the paired difference is not significant), instead of about six times larger. To see why, recall that with $10$ queries, the learner places every boundary point at the middle of its edge. Measured by the error $\max\{\|\tilde u-\hat u_i\|,|\tilde\tau-\hat\tau_i|\}$, the boundaries learned from $10$ queries are about as accurate in both versions of the population (mean error $0.26$ with the offsets of Table~\ref{tab:finite-query-setup} and $0.24$ with the offsets above), and in both cases more accurate than those learned from $20$ queries (mean error $0.33$ and $0.50$), since a single step of binary search can move a boundary point further from the truth than the middle of its edge. What differs is the direction of the errors. With the offsets of Table~\ref{tab:finite-query-setup}, most thresholds in the conflicting-blocks population lie well below the middle of the utility range, so the boundaries learned from $10$ queries make the sampled agents look stricter: at the full-information optimum, their learned violations exceed their true violations by $0.135$ on average, compared with $0.014$ with the offsets above. Errors in the same direction shift the computed lottery, while errors in different directions partly cancel. Since the guarantee of Theorem~\ref{thm:alg_error_bound} only depends on the size of the errors, it does not capture this difference.

\begin{figure}[t]
  \centering
  \includegraphics[width=.55\linewidth]{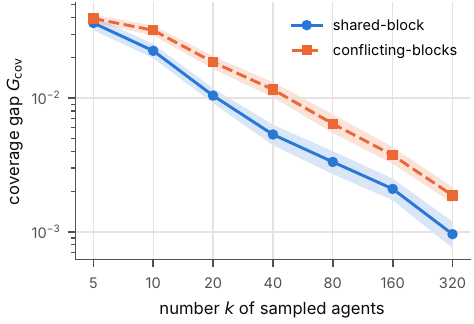}
  \caption{\textbf{Coverage gap with exact recovery.} Every sampled agent receives $109$ queries and is learned exactly, so the total excess equals the coverage gap $G_{\mathrm{cov}}$. Lines show means over 100 populations, and shaded bands are 95\% bootstrap confidence intervals for the means; both axes are logarithmic.}
  \label{fig:appendix-query-diagnostics}
\end{figure}

\paragraph{Exact recovery for every sampled agent.}
To show the coverage gap on its own, we also ran the algorithm with $109$ queries for every sampled agent, for the same values of $k$ and the same populations. Every sampled agent is then learned exactly, and the boundary-learning gap is below $10^{-12}$ in every run. Figure~\ref{fig:appendix-query-diagnostics} shows that the coverage gap decreases steadily with $k$: least-squares fits to the mean curves on a log-log scale have slopes $-0.87$ for the shared-block population and $-0.75$ for the conflicting-blocks population.

\paragraph{Power-mean sweep.}
The $p$-sweep experiment examines the role of the parameter $p$. Unlike the finite-query experiments, it uses one structured instance, with $n=300$ agents and $m=6$ alternatives, and makes no queries. The agents form three clusters of sizes $180$, $75$, and $45$, which we call majority easy, minority hard, and orthogonal small. The agents are generated as in the finite-query experiments, without quantization, from the cluster centers $(2.0, 0.85, 0.10, -0.95, -0.95, -1.05)$, $(-0.95, 2.45, -0.65, -0.35, -0.25, -0.25)$, and $(-0.50, -0.60, 1.80, 0.75, -0.70, -0.75)$, with agent noise of standard deviation $0.08$, $0.07$, and $0.08$, threshold offsets $0.52$, $0.84$, and $0.50$, threshold noise of standard deviation $0.025$, and thresholds clipped to $[0.35,0.88]$. For each $p\in\{1,1.25,1.5,2,3,5,8,12,20,\infty\}$, we compute a minimizer $\bx_p$ of $V_p$ and report the average violation $V_1(\bx_p)$ and the largest violation $V_\infty(\bx_p)$.

Small $p$ emphasizes the average violation, and large $p$ emphasizes the largest violations. In this instance, $V_1(\bx_p)$ increases and $V_\infty(\bx_p)$ decreases as $p$ grows. Moving from $p=1$ to $p=\infty$ lowers the largest violation by 42\%, from $0.391$ to $0.225$, while raising the average violation by 42\%, from $0.047$ to $0.066$.

\begin{figure}[t]
  \centering
  \includegraphics[width=\linewidth]{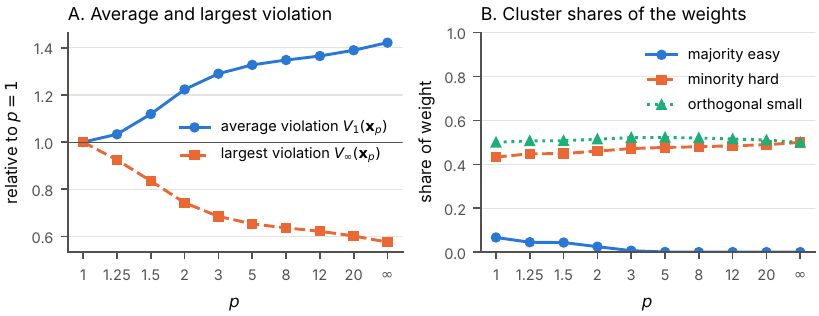}
  \caption{\textbf{Changing $p$ changes which agents drive the optimum.} Panel A plots the average and the largest violation of the minimizers $\bx_p$, relative to their values for $p=1$: larger $p$ lowers the largest violation $V_\infty(\bx_p)$ but raises the average violation $V_1(\bx_p)$. Panel B shows the share of each cluster in the weights $v_i(\bx_p)^{p-1}$ of Theorem~\ref{thm:dual-finite-p} for $1<p<\infty$. The endpoint shares at $p=1$ and $p=\infty$ use the descriptive conventions below and need not be dual-optimal weights.}
  \label{fig:p-sweep}
\end{figure}

\begin{figure}[t]
  \centering
  \includegraphics[width=\linewidth]{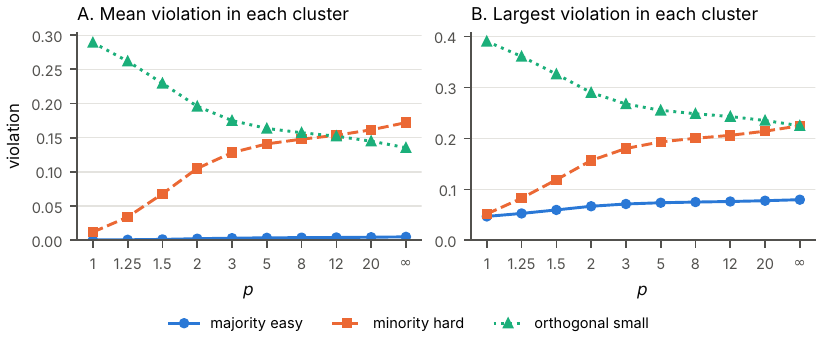}
  \caption{\textbf{Cluster-level effects of changing $p$.} Panel A shows the mean violation within each cluster, and Panel B shows the largest violation within each cluster.}
  \label{fig:appendix-p-clusters}
\end{figure}

For $1<p<\infty$, Theorem~\ref{thm:dual-finite-p} gives agent $i$ the weight $v_i(\bx_p)^{p-1}$. We normalize these weights to sum to $1$, and define the share of a cluster as the sum of the normalized weights of its agents. As $p$ increases, the share of the majority easy cluster, whose agents have the smallest violations, drops from $0.07$ to $0$, and its weight moves to the two clusters with larger violations. For $p=1$, we give equal weight to all agents with violation above $10^{-8}$, and for $p=\infty$, we give equal weight to all agents whose violation is at least $\max_j v_j(\bx_\infty)-\max\{10^{-8},\,10^{-5}\max_j v_j(\bx_\infty)\}$. For $1<p<\infty$, we also checked the condition of Theorem~\ref{thm:dual-finite-p} by comparing the weighted scores of the alternatives used by $\bx_p$ (those with probability above $10^{-6}$) with the largest weighted score; the largest relative difference is $8.5\times 10^{-7}$. Figure~\ref{fig:appendix-p-clusters} shows the violations within each cluster: larger $p$ protects the agents with the largest violations, in the orthogonal small cluster, but shifts some of the violation to the minority hard cluster.

\paragraph{Seeds and confidence intervals.}
Population $s\in\{0,1,\dots,99\}$ of the shared-block and conflicting-blocks families is generated with the NumPy random generator seeded by $(1,s)$ and $(2,s)$, respectively. The sample of $k$ agents from population $s$ of family $f\in\{1,2\}$ is drawn with seed $(f,s,k,e)$, where $e=1$ for the fixed-budget experiment, $e=2$ for the experiment with exact recovery, and $e=3$ for the experiment with thresholds closer to the middle. The instance of the $p$-sweep uses seed $3301$. Confidence intervals are 95\% percentile bootstrap intervals for the mean over the $100$ populations, with $2000$ resamples drawn with seed $2026$.

\paragraph{Compute resources.}
All experiments were run on a Linux machine with $2$ CPU cores (Intel Xeon, 2.8\,GHz) and $8$\,GB of memory, without a GPU. Running all experiments, including the tests of the code, takes less than five minutes. Pilot runs, including those used to choose the population parameters, used a comparable amount of compute.

\paragraph{Limitations.}
The populations are synthetic, and their parameters were chosen so that the effects studied in the theory are visible; as the experiment with thresholds closer to the middle shows, the shape of the curves depends on these choices. With few queries per agent, the learner is the procedure of Lemma~\ref{lem:learn-halfspace} stopped early, and other learners may behave differently. The simulations do not model strategic or inconsistent answers. Finally, the full population is used to evaluate the computed lotteries, which lets us compare them with the full-information optimum but is not possible in deployment. The experiments therefore make a narrow claim: on transparent instances, the effects of a limited budget and of the parameter $p$ are those identified by the theory.

\end{document}